\pdfoutput=1

\documentclass[11pt]{article}

\usepackage[]{acl}

\usepackage{times}
\usepackage{latexsym}

\usepackage[T1]{fontenc}

\usepackage[utf8]{inputenc}

\usepackage{microtype}

\usepackage{inconsolata}

\usepackage{url}

\title{Evaluating Large Language Models as Generative User Simulators for Conversational Recommendation}

\author{
  Se-eun Yoon \hspace{0.5cm} Zhankui He \hspace{0.5cm} Jessica Maria Echterhoff \hspace{0.5cm} Julian McAuley \\
  University of California, San Diego \\
  \texttt{\{seeuny, zhh004, jechterh, jmcauley\}@ucsd.edu}
}

\usepackage{graphicx}
\usepackage{subcaption}
\usepackage{multirow}
\usepackage{makecell}
\usepackage{booktabs}
\usepackage{mdframed}
\usepackage{tabularx}
\usepackage{array}
\usepackage{enumitem}
\usepackage{verbatim}

\newcommand{\taskone}{ItemsTalk}
\newcommand{\tasktwo}{BinPref}
\newcommand{\taskthree}{OpenPref}
\newcommand{\taskfour}{RecRequest}
\newcommand{\taskfive}{Feedback}

\newcommand{\vanilla}{Vanilla LLM}
\newcommand{\demographic}{DI}
\newcommand{\pickiness}{DI + PP}
\newcommand{\interaction}{IH}

\newcommand{\boldheading}[1]{%
    \vspace{0.5em} 
    \noindent\textbf{#1}\hspace{0.1em} 
}

\newcommand{\noboldheading}[1]{%
    \vspace{0.5em} 
    \noindent\textit{#1}\hspace{0.1em}
}
\newcommand{\finding}[1]{%
    \vspace{0.5em} 
    \noindent\textbf{#1} 
}
\newcommand\mtitle[1]{\textcolor{orange}{#1}}
\newcommand\feature[1]{\textcolor{teal}{#1}}

\begin{document}
\maketitle
\begin{abstract}
Synthetic users are cost-effective proxies for real users in the evaluation of conversational recommender systems. 
Large language models show promise in simulating human-like behavior, raising the question of their ability to represent a diverse population of users.
We introduce a new protocol to measure the degree to which language models can accurately emulate human behavior in conversational recommendation. 
This protocol is comprised of five tasks, each designed to evaluate a key property that a synthetic user should exhibit: choosing which items to talk about, expressing binary preferences, expressing open-ended preferences, requesting recommendations, and giving feedback.
Through evaluation of baseline simulators, we demonstrate these tasks effectively reveal deviations of language models from human behavior, and offer insights on how to reduce the deviations with model selection and prompting strategies.\footnote{We release our code and datasets at \url{https://github.com/granelle/naacl24-user-sim}.}
\end{abstract}

\section{Introduction}
In everyday life, recommendations are often sought through conversations: we ask others for advice on which movies to watch, appliances to buy, or restaurants to explore. 
Such experience is what conversational recommendation systems (CRSs) seek to provide, by developing autonomous agents that could chat with users, understand their needs, and provide well-tailored recommendations.
A core challenge that hinders the advancement of the field is  evaluation~\cite{gao2021advances}.
While an ideal approach would involve comprehensive testing with real user interactions, the associated costs and risks drive studies towards proxy methods, which are limited in representing real user evaluation. 
Offline evaluation restricts evaluation to non-interactive modes, allowing only single-turn assessments~\cite{li2018towards, moon2019opendialkg, he2023large}. 
To enable interactive evaluation, studies have introduced synthetic users.
However, they are overly simplified representations of human users, being restricted to binary responses (e.g., yes or no)~\cite{christakopoulou2016towards, lei2020estimation} or holding `target' items as if users and agents are playing guessing games~\cite{sun2018conversational, lei2020interactive, guo2018dialog}.
Other line of work restrict interactions to predetermined rules and templates~\cite{zhang2020evaluating, zhang2022analyzing}.
Essentially, these user simulators suffer from an inherent constraint: they are static (i.e., confined to a finite set of actions), not generative.

\begin{figure}
    \centering
    \includegraphics[width=0.85\linewidth]{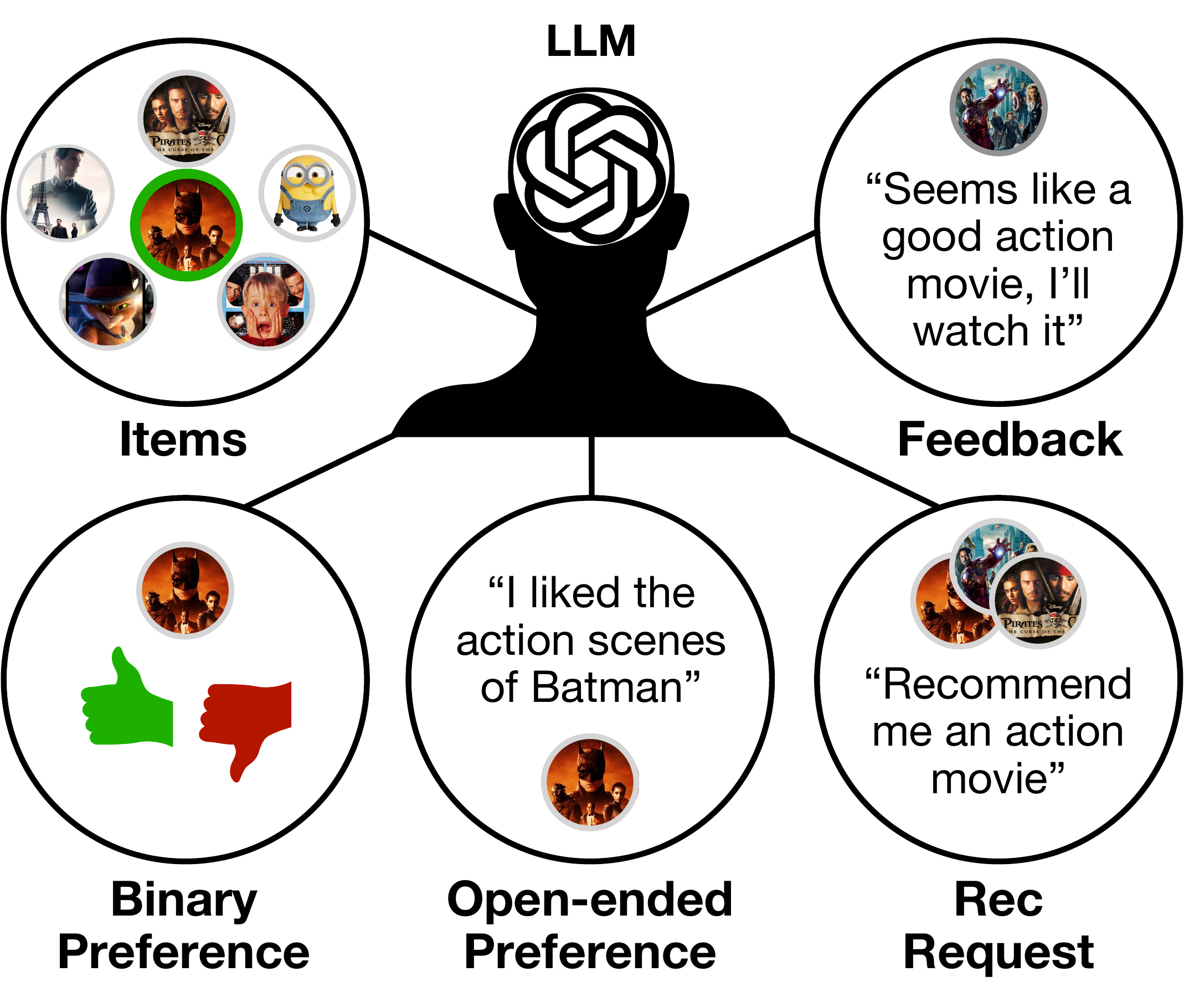}
    \caption{To be successful user simulators for conversational recommendation, representing a population of users, LLMs must fulfill a variety of tasks.}
    \label{fig:fig1}
\vspace{-0.3cm}
\end{figure}

Recently, large language models (LLMs) have demonstrated impressive proficiency in conversational tasks~\cite{pan2023preliminary, zhao2023chatgpt}, motivating a growing number of works to explore their capacity to simulate human behavior~\cite{park2023generative, argyle2023out, aher2023using, gao2023s, momennejad2023evaluating}.
Agents simulated by LLMs are \emph{generative}; 
conditioned upon profiles and memories, these agents exhibit emergent behaviors that appear believable~\cite{park2023generative, qian2023communicative, gao2023s}.
Studies have also explored the use of LLMs as user simulators for recommender systems~\cite{wang2023recagent, wang2023rethinking}.
An important question in each of these studies is to evaluate how closely these simulators represent humans in the task. 
While there are automatic evaluation protocols for replicating general human behavior~\cite{aher2023using, momennejad2023evaluating}, no protocol exists in the context of recommendation.

In conversational recommendation, the requirements of user simulators are distinct from general-purpose human simulators.
The goal is to simulate a \textit{population} of users, each with distinct \textit{preferences}, in a way that these preferences collectively reflect the characteristics of human preferences.
Real user preferences are highly granular and diverse, shaped by each individual's particular set of traits, interaction history, and circumstances.
Such uniqueness is reflected in conversational utterances among individuals, each mentioning distinct items, expressing various preferences, and making highly personalized recommendation requests.
There are also population-level patterns in preferences, such as users preferring some items over others.
Protocols in other domains are unsuitable for evaluating the requirements in conversational recommendation, since they do not consider the behaviors driven by personal preferences \cite{momennejad2023evaluating, aher2023using}.
Although some work considers the uniqueness of simulated individuals, evaluations are limited to manual case studies~\cite{wang2023recagent, park2023generative}.

We propose a new evaluation protocol for measuring the extent to which LLM-based simulators can represent users in conversational recommendation.
This objective poses new challenges:
First, we lack data that maps inputs to ground-truth outputs, such as demographics to survey outcomes~\cite{santurkar2023whose}. 
Second, our outcomes are free-form text, unlike previous work where behavioral outcomes are are choices or numerical values for ground truth comparison~\cite{aher2023using}.
Third, the infinite possibilities of conversational trajectories make the concept of `ground truth' increasingly ambiguous as conversations unfold.

We tackle these challenges by decomposing evaluation into five independent tasks, each measuring a key property that a user simulator should exhibit. 
Each task prompts a simulator and stores the outcomes of a population of simulators.
The outcomes can then be compared to the human data we curate from four different platforms.
The tasks by themselves do not guarantee simulators to be perfect representations of human users, but rather, capture \textit{distortions} in simulators, which is the systematic difference from humans~\cite{aher2023using}.

We demonstrate the effectiveness of these tasks by applying them to baseline simulators and revealing the distortions present in these simulators.
We observe that simulators tend to favor mentioning popular items, correlate little with human preferences, exhibit lack of personalization in requests, and occasionally give incoherent feedback.
We also identify methods to reduce the gap between simulators and humans, indicating that our evaluation protocol can guide future research in developing more realistic user simulators.

Our work is summarized as follows:
\begin{itemize}
\setlength\itemsep{-0.2em}
    \item We propose the first evaluation protocol for LLM-based user simulation in conversational recommendation. Our protocol allows automatic and reproducible evaluation through five tasks and real user datasets.
    \item By running our tasks, we show how simulators could differ from real users. Discrepancies include low item diversity, low correlation with human preference, lack of request personalization, and incoherent feedback.
    \item We show that the gaps can be reduced through prompting and model selection strategies.
\end{itemize}

\begin{table*}[t!]
\centering
\scalebox{0.9}{
\begin{tabularx}{1.1\textwidth}{l|l|l|X}
\hline
\bf Task & \bf Baselines & \bf Datasets & \bf Example prompt \\ 
\hline
\hline
(T1) \taskone & \demographic, \interaction  & ReDial, Reddit, IMDB & 
\small A person mentions Concussion (2015) and Jerry Maguire (1996) in a conversation about movies and proceeds to mention 2 more. What would these 2 movies be? \\
\hline
(T2) \tasktwo & \demographic, \pickiness & MovieLens & 
\small Pretend to be Ms. Guzman. You watched the movie Whiplash (2014). Did you like the movie? Answer Yes or No. Don't say anything else.\\ 
\hline
(T3) \taskthree  & \demographic, \pickiness & IMDB & 
\small Pretend to be Mr. Li. You watched the movie The Bellboy (1960). What are your thoughts on this movie? Answer should not exceed 809 characters. \\ 
\hline
(T4) \taskfour & \vanilla & Reddit & 
\small Generate a movie recommendation request. Include the following movies in your text: Oldboy (2003), Memento (2000). Length of the request is approximately 374 characters.\\ 
\hline
(T5) \taskfive & \vanilla & Reddit & 
\small In the following conversation ...
    If the recommendation is coherent to your request, answer Accept.
    If the recommendation is incoherent to your request, answer Reject.\\ 
\hline
\end{tabularx}
}
\caption{Tasks overview. Prompts are partially displayed. See \ref{sec:prompts} for full prompt descriptions. (DI: \underline{D}emographic Information, IH: \underline{I}nteraction \underline{H}istory, PP: \underline{P}ickiness \underline{P}ersonality)}
\label{tab:overview}
\end{table*}

\section{Evaluation Tasks}

Here we first introduce our tasks and later explain the execution of these tasks in Section~\ref{sec:methods}.

\noboldheading{(T1) \textbf{\taskone}: Choosing items to talk about.} 
Often when users talk about recommendations, they mention items.
Contexts may vary: to request similar items, to express preference on certain items, or to simply chat about an item~\cite{li2018towards}.
We compare the distributions of items mentioned by simulators and real users.

\noboldheading{(T2) \textbf{\tasktwo}: Expressing binary preference.} 
Binary questions, such as `Did you enjoy this movie?' are  commonly observed in conversations~\cite{li2018towards}. 
While answers need not be binary, we fix the answers to binary in this task and examine how well simulators reflect human preferences.

\noboldheading{(T3) \textbf{\taskthree}: Expressing open-ended preference.} 
Open-ended utterances allow users to express detailed preferences, such as appreciating the cast of a movie while finding the plot uneventful~\cite{xia2023user}. 
We examine whether simulators can express preferences on aspects of items (e.g., cast and plot), and whether the aspects and preferences are similar to those expressed by real users.

\noboldheading{(T4) \textbf{\taskfour}: Requesting recommendations.} 
A need for recommendation is verbalized through requests.
Requests can range from something general, such as `Recommend me a good movie,' to a more personalized demand, such as `Recommend me a movie that involves a lawyer or a magician but does not contain action scenes.'
While related to preferences, requests stem from immediate demand, such as being in a mood for a certain movie.
Given the vastness of tastes and circumstances, a wide variety of requests may emerge~\cite{he2023large}.
We investigate whether requests generated by simulators are as diverse as those of real users.

\noboldheading{(T5) \textbf{\taskfive}: Giving feedback.} 
To evaluate CRSs, simulators should be able to provide final feedback of whether the recommendation was successful~\cite{wang2023rethinking}.
(Real users may or may not provide explicit feedback, but they have a general impression of whether the recommendation was satisfactory.)
Particularly, if recommendations and explanations are relevant to one's requests and preferences, one should be likely to accept the recommendation. 
If irrelevant, one should reject them.
We examine whether simulators can exhibit such coherent patterns of feedback generation.
\section{Methods}
\label{sec:methods}

Our protocol treats the design choices of simulators as a black box.
We only require that simulators should accept free-form natural language as input and generate language as output.
As noted in the introduction, we consider the population of users, since recommender systems are tested against a large group of users. 
Importantly, our focus is not to replicate a fixed pool of users, but to generate a \textit{new} group of users whose behavior characteristics resemble those of human users.
Tasks should be zero-shot; simulators should not be trained or conditioned on our tasks, nor be informed about our evaluation metrics.
This is to avoid simulators fitting to the tasks instead of performing well in generic situations.

\boldheading{Datasets}
We use real-world datasets to compare simulator outputs to human output. Dataset statistics are summarized in \ref{sec:dataset_statistics}.
\textbf{ReDial}~\cite{li2018towards} consists of multi-turn conversations, where one person plays the role of a movie seeker, and the other as a recommender. We use the seeker side of this dataset.\footnote{The Mechanical Turk workers may differ from real users with genuine incentives. However, we assume the person playing as seeker bears insignificant difference from the real seeker, as the role requires less effort and expertise.}
\textbf{Reddit}~\cite{he2023large} consists of conversations in Reddit communities on movie recommendations. Users post requests for recommendations and other users comment on this post with movies, sometimes with explanations. 
\textbf{MovieLens}~\cite{harper2015movielens} is a movie ratings dataset consisting of 25M ratings. 
\textbf{IMDB} is a movie review dataset from IMDB,\footnote{\url{https://www.imdb.com/}}, aggregated per user. 
Each task uses different dataset(s), since the datasets are heterogeneous and are not applicable to every task (see Table~\ref{tab:overview}). 
We only include movies up to the year 2021 to ensure that LLMs have knowledge about the movies.

\boldheading{Baselines}
We use prompt-based simulators as baselines, using OpenAI~\cite{openai2021} models \textit{gpt-3.5-turbo}, \textit{gpt-4}, and \textit{text-davinci-003}. 
The baselines can be used in any task, but for each task, we select the baselines that give the best insights (see Table~\ref{tab:overview}).
\textbf{\vanilla} runs without any specialized prompts designed to induce variability in outputs. 
Instead, it relies solely on the inherent variability of LLM outputs.\footnote{The degree of variability can be adjusted by the temperature parameter, but we use the default temperature values across all LLMs.}
\textbf{\demographic} (Demographic Information) is a method used by \citet{aher2023using} to simulate gender and racial diversity. 
We follow their method and sample from titles Mr. and Ms.,\footnote{There are more titles, but we use these two for simplicity.} and 500 most common surnames across five racial groups.\footnote{\url{https://www.census.gov/topics/population/genealogy/data/2010_surnames.html}}
\textbf{\pickiness} (Pickiness Personality) adds a personality trait to demographic information, that is, pickiness toward movies.
For each simulator, we randomly sample one of three pickiness levels: not picky, moderately picky, and extremely picky.
\textbf{\interaction} (Interaction History) samples a set of interactions from a real user and prompts to act like this user. 
The interaction may be a subset of mentioned items (ReDial), mentioned items with timestamps (Reddit), or reviews of items (IMDB).

\boldheading{Execution and evaluation}
For each task, we create a population of simulators, each given a task-specific prompt. 
Example prompts are in Table~\ref{tab:overview} and all the prompts are in \ref{sec:prompts}.

\textit{\taskone} prompts the simulator to mention a certain number of items.
Each prompt uses a single dataset entry to determine the number of items to mention and interaction history (for the {\interaction} baseline); the number of prompts equals the number of dataset entries.
For evaluation, we compare the distribution of mentioned items between the simulator and the dataset (items in prompt are removed). 
The diversity of distribution is summarized by entropy: \( H(X) = -\sum_{i=1}^{n} p(x_i) \log p(x_i) \), where \( p(x_i) \) is the probability that an item \( x_i \) is mentioned.

\textit{\tasktwo} prompts the simulator to act as if one has interacted with an item, and asks whether one has positive opinions on it.
We sample two groups of $200$ movies from MovieLens: frequent ($\geq5000$ ratings) and infrequent ($\leq500$ ratings). 
This is to observe if the simulators reflect human preferences better on frequent movies. 
The distribution of average rating ranges from 1 to 5.
For each movie, we run $100$ simulators to output a binary preference and get the proportion of `Yes' answers (i.e., positive rate). 
We compute the Pearson correlation coefficient between the average rating and positive rate. 

\textit{\taskthree} prompts the simulator to assume one has interacted with a certain item, and asks one's thoughts on it. 
Each prompt uses a review from IMDB to obtain target response length. 
After getting a collection of responses, we conduct aspect-based sentiment analysis with PyABSA~\cite{yang2023pyabsa}. 
We compare the aspect and sentiment distributions of humans and simulators. 

\textit{\taskfour} prompts the simulator to generate a recommendation request containing a set of items. 
The reason we include the set of items is to evaluate only the capability to generate requests, and not items included in the request.
In each prompt, the items and target length are determined by a real user request in the Reddit dataset; we obtain the same number of requests.
We compare the diversity and granularity of synthetic and real requests by computing type-token ratio, word embeddings with Word2Vec~\cite{mikolov2013efficient}, and sentence (request) embeddings with SBERT~\cite{reimers2019sentence}.
We use the cosine diversity of embeddings~\cite{anderson2020algorithmic}: 
$$
1 - \frac{1}{N} \sum_{i=1}^N \frac{\vec{s}_i \cdot \hat{\mu}}{\|\vec{s}_i\| \cdot \|\hat{\mu}\|}    
$$
where $N$ is the number of embeddings, $\vec{s}_i$ is the $i^{th}$ embedding, and $\hat{\mu}$ is the centroid $\hat{\mu} = \sum_i \vec{s}_i / N$.

\textit{\taskfive} prompts the simulator to provide feedback to request-recommendation pairs. 
For each request in the Reddit dataset, we sample: (1) a comment from this request (positive recommendation) (2) a random comment (negative recommendation).
We formulate two sub-tasks. 
\textit{Accept/reject:} simulators should reject negative recommendations. 
\textit{Comparison:} simulators should prefer positive recommendations over negative ones. 

\begin{table}[t] 
\centering 
\caption{Entropy of mentioned items. Simulators yield lower entropy, indicating lower diversity. Prompting with interaction history enhances diversity.}
\scalebox{0.9}{
\begin{tabular}{c|ccc}
\hline
Generator & IMDB & Reddit & ReDial \\
\hline
\hline
Human & \bf 12.61 & \bf  11.73 & \bf  9.71 \\
\hline
\multicolumn{4}{c}{Demographic information} \\ 
\hline
gpt-3.5 & 4.79 & 3.97	& 4.00 \\ 
gpt-4 & 5.29 & 4.78 & 4.18 \\
text-davinci & 6.42 & 6.69 & 6.66 \\
\hline
\multicolumn{4}{c}{Interaction history} \\ 
\hline
gpt-3.5 & 7.96 & 7.14 & 7.68 \\
gpt-4 & 8.59 & 9.50 & 9.03 \\
text-davinci & 10.79 & 9.97 & 8.63 \\
\hline
\end{tabular}
}
\label{tab:items}
\end{table}
\begin{figure}[t]
    \centering
    \includegraphics[width=0.49\textwidth]{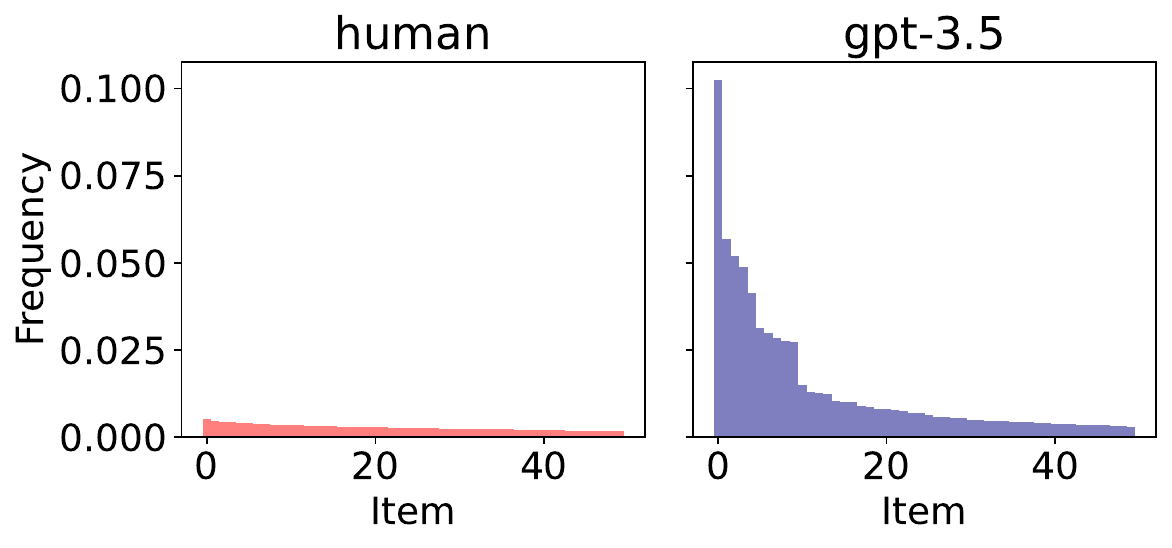}
    \vspace{-2em}
    \caption{Distribution of mentioned items (Reddit+\interaction). Items are sorted in descending frequency. Humans mention more diverse items (left) than simulators (right).}
    \label{fig:items}
\end{figure}
\begin{figure}[t]
    \centering
    \includegraphics[width=0.5\textwidth]{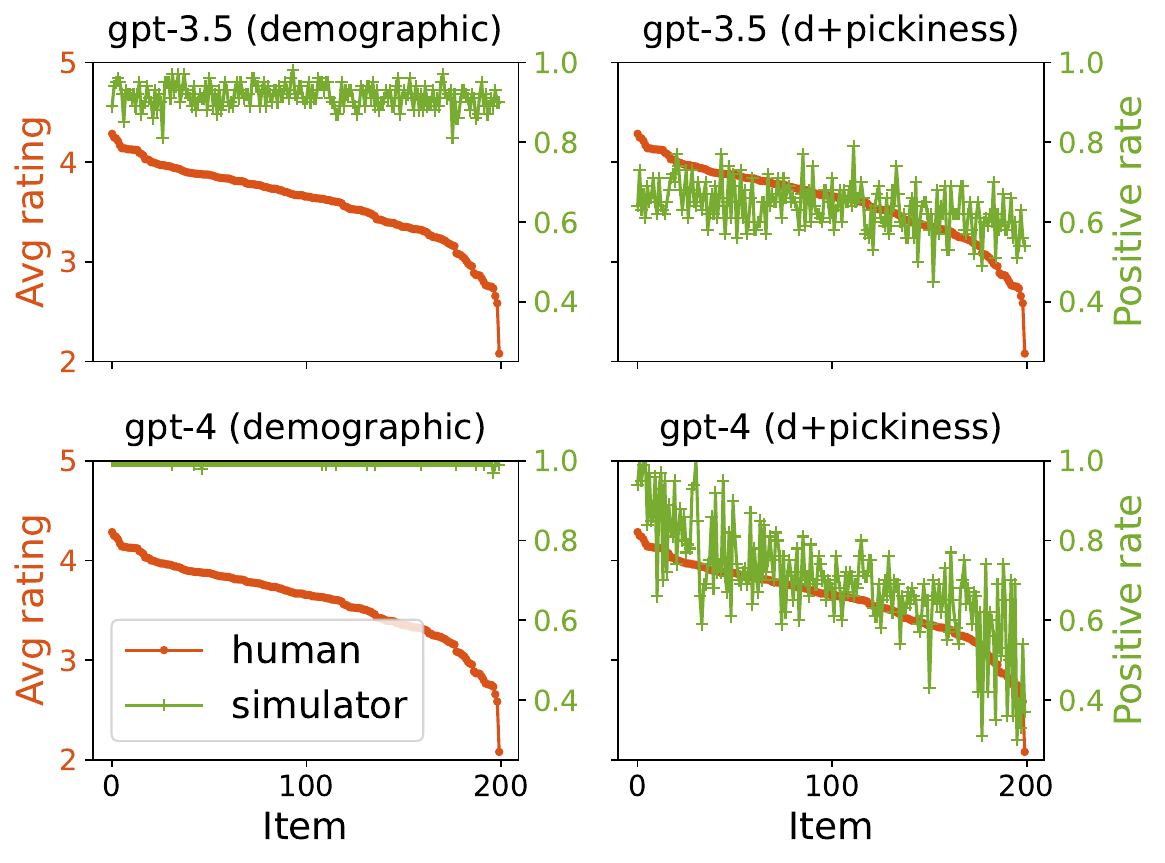}
    \caption{How well do simulators reflect human preferences? Most fail, except gpt-4 with pickiness (bottom right). The units for ratings and positive rates are different but included in the same plot to compare trends.}
    \label{fig:bin}
\end{figure}
\begin{table}[t] 
\centering
\caption{Correlation coefficient between human and simulator preferences. Higher correlation is better, showing the effect of providing pickiness personality. `Undefined' indicate undefined correlation; all simulators responded `yes'. P-values are less than $0.05$.}
\label{tab:correlation}
\scalebox{0.9}{
\begin{tabular}{c|cc}
\hline
Generator & Frequent items & Infrequent items \\
\hline
\hline
\multicolumn{3}{c}{Demographic information} \\ 
\hline
gpt-3.5 & 0.18 & 0.12 \\ 
gpt-4 &  0.24 & 0.53 \\
text-davinci & Undefined & 0.29 \\
\hline
\multicolumn{3}{c}{Demographic information + Pickiness} \\ 
\hline
gpt-3.5 & 0.45 & 0.36 \\
gpt-4 & 0.75 & 0.76 \\
text-davinci & 0.49 & 0.64 \\
\hline
\end{tabular}
}
\end{table}
\begin{figure}[t]
    \centering
    \includegraphics[width=0.49\textwidth]{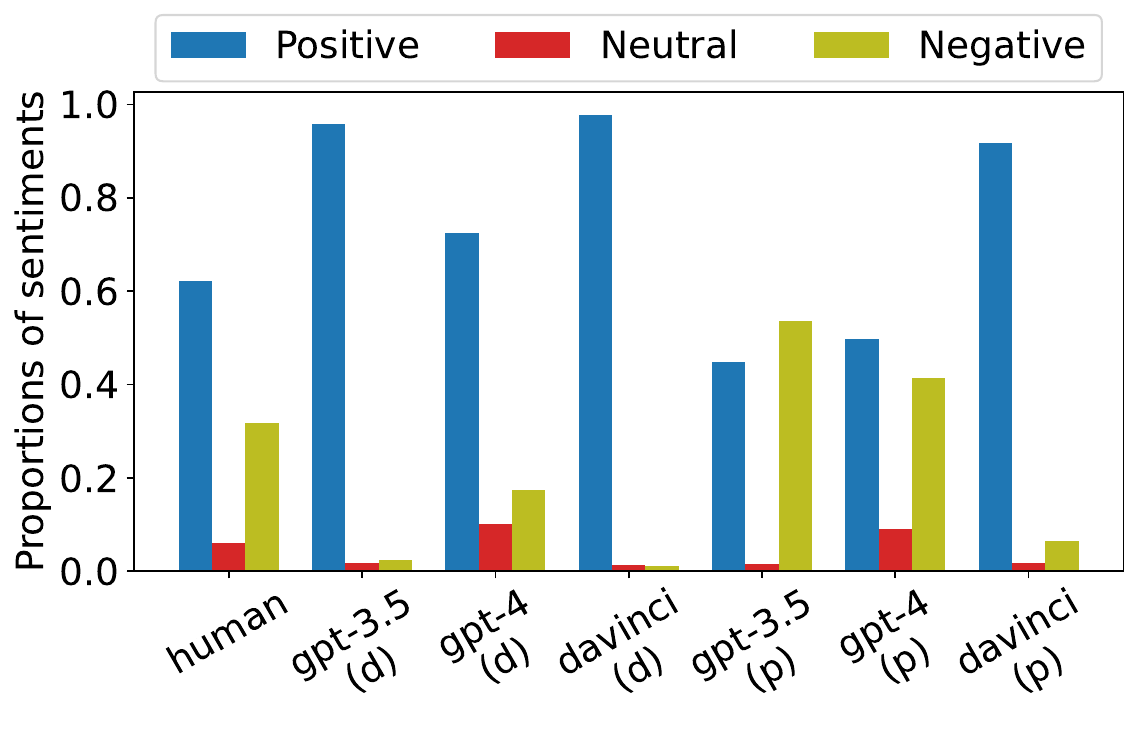}
    \vspace{-2em}
    \caption{Sentiments in open-ended responses.}
    \label{fig:sentiments}
\end{figure}
\begin{table}
\caption{Aspects and sentiments in open-ended responses. Humans have low number of aspects but high aspect entropy and sentiment entropy.}
\scalebox{0.9}{
\begin{tabular}{c|ccc}
\hline
Generator & \# aspects & \makecell[c]{Aspect\\entropy} & \makecell[c]{Sentiment\\entropy} \\
\hline
\hline
Human & 85 & \bf 5.85 & 1.19 \\
\hline
\multicolumn{4}{c}{Demographic information} \\ 
\hline
gpt-3.5 & 71 & 4.86	& 0.29 \\ 
gpt-4 & 97 & 5.57 & 1.11 \\
text-davinci & 194 & 5.63 & 0.18 \\
\hline
\multicolumn{4}{c}{Demographic information + Pickiness} \\ 
\hline
gpt-3.5 & 101 & 5.20 & 1.09 \\
gpt-4 & 97 & 5.59 & \bf 1.34 \\
text-davinci & \bf 232 & 5.47 & 0.48 \\
\hline
\end{tabular}
}
\label{tab:aspects}
\end{table}

\section{Experiments}

We summarize our experiment results as follows.

\finding{Finding 1: Simulators mention less diverse items compared to real users.} 
Our first task, \textit{\taskone}, reveals that the distribution of items mentioned by simulators is heavily skewed toward popular items, in contrast to a more even distribution of items mentioned by humans (Figure~\ref{fig:items}).
We observe the trend across all baselines and datasets, quantified by entropy (Table~\ref{tab:items}). 

\finding{Finding 2: Prompting with interaction history enhances item diversity.}
Comfortingly, prompting with interaction history yields much higher diversity than prompting with demographic information, and we even observe cases (gpt-4 and text-davinci-003) prompted with interaction history from Reddit and IMDB slightly exceed the diversity of humans in ReDial.
This suggests that interaction history (`trigger' items) is a strong condition for generating diverse items.

\finding{Finding 3: Simulators may poorly represent real user preferences.}
Our next task, \textit{\tasktwo}, captures simulators failing to represent human preferences.
In Figure~\ref{fig:bin}, we sort items in decreasing average rating and plot the positive rate (proportion of simulators that answered `yes' to whether they liked the movie).
Positive rates remain constant regardless of human preferences, except gpt-4 + \pickiness, where the positive rate decreases as average rating decreases. 
Unexpectedly, higher item frequency (how well known is an item, measured by number of ratings) does not necessarily lead to better preference alignment (Table~\ref{tab:correlation}), despite LLMs likely being more exposed to these items during training. 
We show all the results in \ref{sec:more_tasktwo}.

\finding{Finding 4: Adding pickiness personality improves preference alignment.}
Endowing simulators with varying levels of pickiness not only diversifies preferences but also improves correlation, sometimes yielding strong correlation (Table~\ref{tab:correlation}). 
This suggests that picky simulators can successfully discern low-rated movies.
Without pickiness, correlations are low to moderate: simulators tend to be consistently optimistic---in one case (text-davinci-003 + \demographic), all the answers were `yes'.

{
\begin{table*}[t]

\centering
\begin{tabularx}{\textwidth}{X|X}
\toprule
\bf Human requests & \bf gpt-3.5-turbo requests \\ 
\midrule
\small Movies showing \feature{extreme loneliness or depression}. I have watched \mtitle{Taxi Driver (1976)} and \mtitle{Joker (2019)} and would like to see more similar movies showing loneliness or depression.
 & \small Looking for a \feature{gripping psychological thriller} similar to \mtitle{Taxi Driver (1976)} or \mtitle{Joker (2019)}? Seeking a movie that delves into the mind of \feature{complex characters}? 
 \\
\midrule
\small Movies about \feature{conspiracies, lies, and finding the truth} like \mtitle{Memento (2000)} and \mtitle{Fight Club (1999)}?. Espcially ones that have \feature{big plot twists}. 
& \small Looking for \feature{mind-bending thrillers} like \mtitle{Fight Club (1999)} and \mtitle{Memento (2000)}. Any suggestions? Need \feature{gripping plots} that \feature{leave me questioning reality}!
\\
\bottomrule
\end{tabularx}
\caption{Examples of recommendation requests. Even when humans and simulators include the same movies (orange), simulators tend to produce more general requests (green), repeating same expressions for different requests. }
\label{tab:request-examples}
\end{table*}
}
\begin{table}[t]
\centering
\caption{Diversity of requests: word diversity (type-token ratio) and embedding (cosine) diversities. Simulators reuse the same words across different requests, generating less personalized requests.}
\scalebox{0.9}{
\begin{tabular}{l|ccc}
\hline 
Generator & Word & \makecell{Word emb.} & \makecell{Sentence emb.} \\ 
\hline
\hline
Human & \bf 0.65 & 0.427 & \bf 0.391 \\
\hline
gpt-3.5 & 0.50 & \bf 0.433 & 0.295 \\
gpt-4 & 0.61 & 0.436 & 0.300 \\
text-davinci & 0.49 & 0.418 & 0.288 \\
\hline
\end{tabular}
}
\label{tab:diversity}
\end{table}

\begin{figure}[t]
    \centering
    \includegraphics[width=0.95\linewidth]{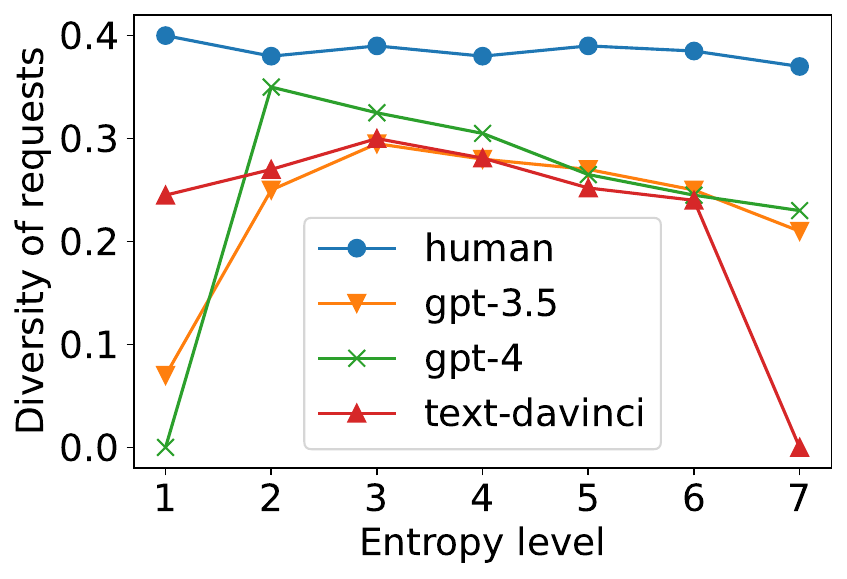}
    \caption{Diversity of requests per entropy level. Simulator requests are less diverse across all entropy levels.}
    \label{fig:requests}
\vspace{-0.3cm}
\end{figure}

\begin{figure}[t]
\centering
\begin{mdframed}
\begin{itemize}[leftmargin=*]
    \item \textbf{Human:} `Movies about alcoholism', `Space movies?', `Movies about redemption', `Inspirational movies', `Good biography movies', `Impactful endings?', `Growth mindset versus Fixed Mindset', `Rock climbing movies', `Movies about nihilism’
    \item \textbf{gpt-3.5-turbo:} `Movie recommendation?', `Need movie suggestions', `Need movie recs!!', `Need movie recommendations', `Movie recs?', `Movie recommendations?'
    \item \textbf{gpt-4:} `Got recs?’
    \item \textbf{text-davinci-003:} `Recommend a movie', `Cheerful movies?', `Recommend me!’
\end{itemize}
\end{mdframed}
\caption{Low-entropy requests generated by humans and simulators.}
\vspace{-0.3cm}
\label{fig:low-entropy}
\end{figure}

\finding{Finding 5: Simulators express preferences differently from real users. Model choice and prompting may mitigate the difference.}
\textit{\taskthree} reveals how simulators express preferences in a way different from humans (Table~\ref{tab:aspects}).
First, simulators generate more sentiment-associated aspects than humans.
Real users often express opinions in subtler ways, such as a movie being `suitable for background noise,' rather than simply praising or criticizing explicit aspects (e.g., cast and plot) of the movie. 
Second, simulators have lower \textit{aspect entropy}, even though they have more aspects. 
This indicates that it is predictable which aspects they will mention, e.g., mentioning the same aspects repeatedly.
Finally, simulators are biased towards positive sentiment, resulting in low sentiment entropy, unless prompted to behave as picky users (Table~\ref{tab:aspects} and Figure~\ref{fig:sentiments}). 
The simulator closest to humans is gpt-4 + \pickiness, with similar aspect and sentiment statistics. 
Therefore, model choice and prompting strategies (e.g., adding pickiness) may enhance realism in simulator preference and expressions of preference.

\finding{Finding 6: Simulators struggle to generate a diverse pool of personalized requests.}
\textit{\taskfour} reveals that simulators generate less personalized requests than real users.
The request diversity, measured by cosine diversity of sentence embeddings, are lower than humans across all models, with gpt-4 (the most diverse) generating $23\%$ less diverse requests than humans (Table~\ref{tab:diversity}).
Simulators have lower diversity across all entropy levels (taking words as variables), particularly in the lowest entropy level (Figure~\ref{fig:requests}). 
For better understanding, we show low-entropy requests in Figure~\ref{fig:low-entropy}, where we see simulators struggling to make specific requests, while humans are specific even when text is short.

We also measure the diversity of words and word embeddings (Table~\ref{tab:diversity}).
Interestingly, simulators have lower \textit{word} diversity and higher \textit{word embedding} diversity:
although simulators generate a semantically diverse range of vocabulary, they tend to \textit{reuse the same words}.
Upon manual inspection, we find that expressions such as `gripping,' `mind-bending,' `compelling,' `keeps me on the edge of my seat' are repeatedly used.
These expressions can be applied to a general range of movies.
Humans, in contrast, tend to express finer-grained preferences, asking for movies that meet a specific criterion.
We show in Table~\ref{tab:request-examples} how the requests differ, even when they contain the same set of movies. 
For example, many movies other than \textit{Taxi Driver (1976)} or \textit{Joker (2019)} can be `gripping psychological thrillers,' but a more limited set of movies deal with `extreme loneliness or depression.' Therefore, the low diversity of requests seems to arise from their \textit{generality}; LLMs make generic requests, while humans are free to be more personal, and hence, diverse in the population level.\footnote{We tried various prompts to generate more diverse requests, but there was no significant difference in the results.}

\begin{figure}[t]
    \centering
    \includegraphics[width=\linewidth]{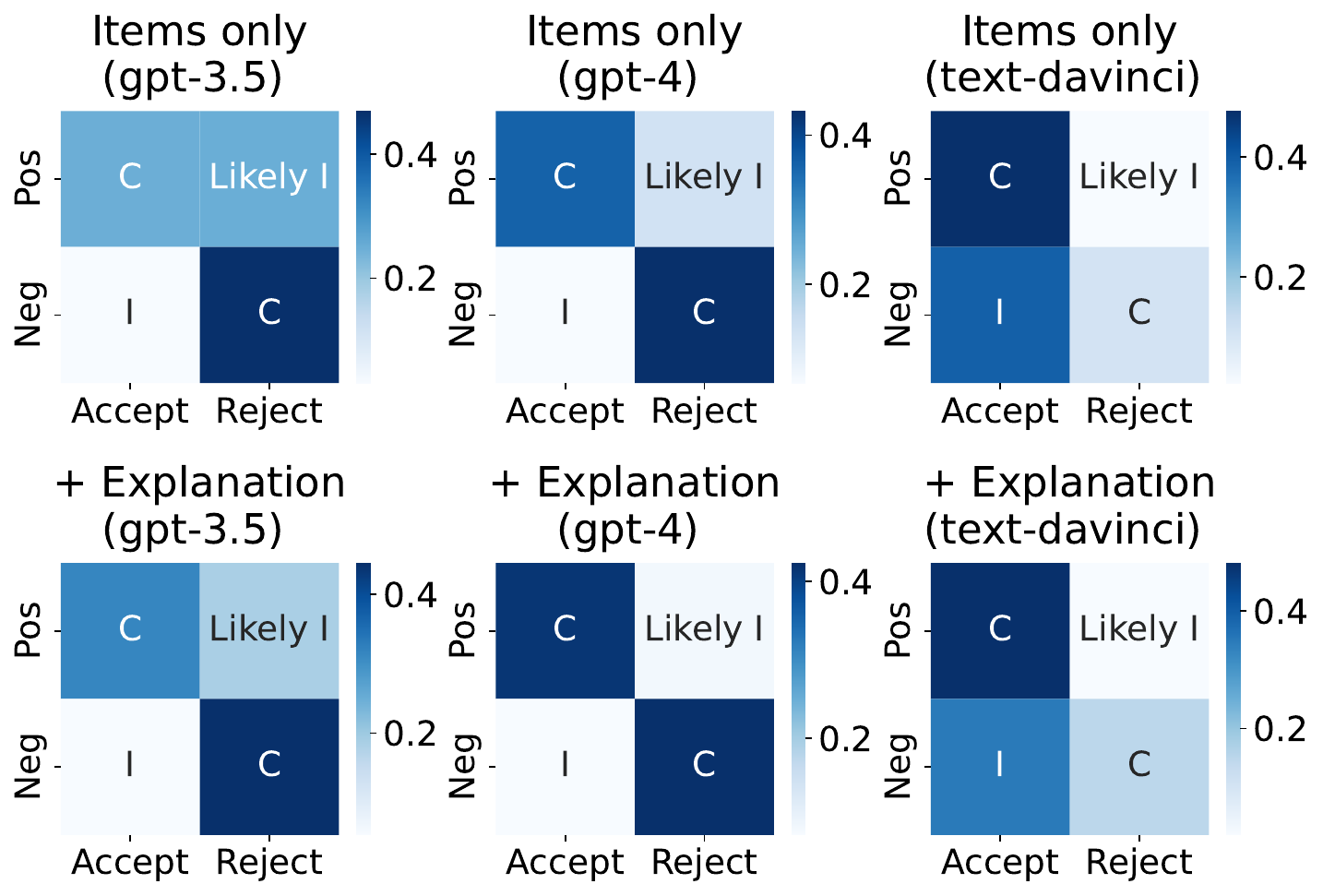}
    \caption{Feedback coherence (accept/reject task). `I' stands for incoherent; `C' stands for coherent. Recommendations contain only items (left column) or items with explanations (right column).}
    \label{fig:feedback}
\vspace{-0.3cm}
\end{figure}

\begin{table}[t]
\centering
\caption{Feedback coherence (proportion of coherent feedback). Simulators are often coherent, but there is room for improvement.}
\scalebox{0.89}{
\begin{tabular}{@{}c|c|cc@{}}
\hline
& Generator & \makecell[c]{Prop. coherent} & \makecell[c]{Prop. neither} \\ 
\hline
\multirow{3}{*}{\makecell[c]{Items\\only}} & gpt-3.5 & 0.8264 & 0.0087   \\
                         & gpt-4                     & 0.9096  & 0.0120  \\ 
                         & text-davinci              & 0.8108  & 0 \\ 
\hline
\multirow{3}{*}{\makecell[c]{Items +\\explain}} & gpt-3.5 & 0.8039 & 0.0049  \\
                       & gpt-4                       & 0.9047  & 0   \\ 
                       & text-davinci          & 0.6567  & 0   \\ 
\hline
\end{tabular}
}
\label{tab:feedback}
\end{table}

\begin{figure}[t]
\centering
\begin{mdframed}
\begin{itemize}[leftmargin=*]
    \item \textbf{Request:} I'm in a state of life rn that i really want/need movies with the Main Character being a loner or alone in general, he doesn't have to be alone the entire movie, but just portrayal of a good loner main character.
    \item \textbf{Recommendation:} Nightcrawler (2014)
    \item \textbf{Feedback from simulator:} Reject.
    \\Nightcrawler is not about a loner main character, but rather about a character who becomes involved in the underground world of crime journalism.
\end{itemize}
\end{mdframed}
\caption{Example feedback from user simulator (gpt-3.5-turbo), given human request and recommendation.}
\label{fig:why}
\vspace{-0.3cm}
\end{figure}

\finding{Finding 7: While simulators often give coherent feedback, there is room for improvement.}
For the accept/reject task, feedback is \textit{coherent} when the simulator rejects negative or accepts positive recommendation. 
Feedback is \textit{incoherent} when the simulator accepts negative recommendation.
The case where the simulator rejects positive recommendation is controversial; a user may still reject a relevant recommendation for reasons external to the request. 
We leave this case as \textit{likely incoherent} and focus evaluation on clearer cases.
As in Figure~\ref{fig:feedback}, simulators are overall coherent, but sometimes give incoherent feedback, its proportion ranging from $3\%$ (gpt-3.5-turbo) to $35\%$ (text-davinci-003).
Particularly, text-davinci-003 is biased towards optimistic feedback, i.e., accepting even if the recommendation is irrelevant.  
We also see that providing explanations encourages simulators accept relevant recommendation.

For the comparison task, feedback is coherent when the simulator prefers positive recommendation over negative recommendation, and incoherent otherwise.
We exclude cases where simulators respond `neither,' and compute its proportion. 
We see in Table~\ref{tab:feedback} that simulators are more often coherent than incoherent, and the proportion of `neither' is negligible. 
Variations exist; coherence ranges from $65\%$ (text-davinci-003) to $90\%$ (gpt-4). 
Interestingly, with explanations, simulators become slightly less coherent.
A possible reason is that explanations add persuasiveness to negative recommendations as well as positive, making it slightly trickier for simulators to distinguish the two.

\textbf{Finding 8: Simulators may not capture subtle nuances in requests, and thus reject relevant recommendations.}
Finally, we ask simulators \textit{why} they have given such feedback.
We manually inspect the responses and find that the reasons are rather compelling, based on accurate factual knowledge.
Incoherence often arise due to missing subtle nuance of what users are asking for.
For instance, in Figure~\ref{fig:why}, the simulator rejects \textit{Nightcrawler (2014)} because the movie is not `about' a loner main character, while the user asks for a movie `with' a loner main character.
More examples are in \ref{sec:more_taskfive}.
We leave a more thorough analysis of the classification of incoherent cases as future work.

\section{Related Work}
\label{sec:related}

\boldheading{User simulation for recommendation}
Early work in conversational recommendation formulate bandit problems to efficiently update traditional models, focusing on item selection and expecting binary preference answers from synthetic users~\cite{christakopoulou2016towards}. 
Recent work considers more realistic conversations with flexibility in natural language, but still confines users to binary or multi-choice responses~\cite{lei2020estimation, lei2020interactive}. 
For evaluation, often a set of target items is predefined per user and the user rejects a recommendation unless the target item is mentioned, and a model is regarded superior if fewer turns are used to reach the target item~\cite{guo2018dialog, sun2018conversational, lei2020estimation, lei2020interactive}.
Alternatively, agenda-based simulation use a state diagram of actions, and recommendation is successful when the conversation reaches the `complete' state~\cite{zhang2020evaluating, zhang2022analyzing}.
However, user actions follow a fixed set of rules and utterance templates, which is unlikely with real users.
Generative simulators, powered by LLMs, effectively avoids this problem, demonstrating more realistic conversation capabilities~\cite{wang2023recagent, wang2023rethinking, zhang2023agentcf}.
However, it is still uncertain how realistic these simulators are compared to real users.

\boldheading{LLMs as human proxies}
There is increasing effort to substitute expensive human experiments with LLMs.
In conversational recommendation, \citet{wang2023rethinking} prompts ChatGPT with target items, which results in users giving `hints' toward specified items. 
\citet{wang2023recagent} goes beyond conversation and introduces a simulation environment where users not only chat about recommendations, but also browse websites, search for items, and share opinions on social media.
Non-verbal user behavior in recommendation is explored in \citet{zhang2023agentcf}.
In other domains, 
\citet{owoicho2023exploiting} and \citet{wang2023depth} explore LLMs as user simulators in conversational search;
\citet{hamalainen2023evaluating} evaluates LLMs for generating synthetic user experience data in HCI; 
\citet{aher2023using} evaluates LLMs for replicating human behavior in social science experiments.
Others create simulation environments where LLM agents interact with each other and generate realistic behaviors~\cite{park2023generative, gao2023s, qian2023communicative}.
No work, to the best of our knowledge, introduces protocols for synthetic users in conversational recommendation.

\boldheading{LLMs for recommendation}
Recent papers explore the use LLMs for recommendation~\cite{hou2023large, li2023gpt4rec, kang2023llms, fan2023recommender, chen2023palr}, but these work focus on LLMs as \textit{recommenders}, not recommendation \textit{seekers}, and are therefore orthogonal to our work.

\section{Conclusion}

We introduce a new protocol for evaluating LLMs as user simulators for conversational recommendation.
We design five evaluation tasks, where each task addresses an essential property for simulators to be realistic user proxies.
By running the tasks on simulators, we show how the tasks effectively reveal discrepancies of simulators from real users. 
Our work aims to set a benchmark for evaluating simulators automatically, with future plans to enhance their realism.

\section*{Limitations} 
Our tasks provide necessary conditions, not sufficient conditions, for simulators to represent a group of real users.  
More tasks could be added to evaluate more properties, such as asking questions about recommendations, or dealing with evolving items (that are not in the training corpus of LLMs). 

While our approach is domain-agnostic, the datasets used in this paper are limited to movies.
Different domains (e.g, e-commerce) may require domain-specific tasks and may produce different results.
An important avenue for future work is to collect CRS datasets in various domains---most existing datasets are on movies or media content. 

Our observations on baseline simulators may not represent all possible simulators. 
In particular, we use OpenAI models, default temperature values, and simple prompt-based baselines. 
More analysis could be made with open-source models, various hyperparameters, and advanced simulators. 

\section*{Ethics Statement}  
Our paper is primarily centered on CRS applications, but has broader implications in making artificial intelligence agents more closely aligned to real humans. 
As simulators exhibit more realistic behavior, the risks of misuse, deception, and over-reliance on these simulators may arise.
A possible way to mitigate these risks is to implement distinct markers on simulators.
For instance, simulators should truthfully disclose that they are not humans.
Finally, we stress that, while simulators are valuable tools for pre-deployment testing, they cannot fully replace human interactions.
Ultimately, real user experiments are needed for final testing.

We use scientific artifacts. 
Redial is licensed under the CC BY 4.0 License. 
MovieLens states that the data may be used for research purposes under a set of conditions (e.g., citation), which this paper meets. 
We ask direct permission from the authors of the Reddit dataset.
IMDB is available for non-commercial usage.
PyABSA is under the MIT License.
Word2Vec and SBERT are under the Apache License, Version 2.0.

\section*{Acknowledgements}
This research was supported by a research grant from Cisco Systems, Inc.

\bibliography{custom}

\begin{thebibliography}{38}
\expandafter\ifx\csname natexlab\endcsname\relax\def\natexlab#1{#1}\fi

\bibitem[{Aher et~al.(2023)Aher, Arriaga, and Kalai}]{aher2023using}
Gati~V Aher, Rosa~I Arriaga, and Adam~Tauman Kalai. 2023.
\newblock Using large language models to simulate multiple humans and replicate
  human subject studies.
\newblock In \emph{ICML}.

\bibitem[{Anderson et~al.(2020)Anderson, Maystre, Anderson, Mehrotra, and
  Lalmas}]{anderson2020algorithmic}
Ashton Anderson, Lucas Maystre, Ian Anderson, Rishabh Mehrotra, and Mounia
  Lalmas. 2020.
\newblock Algorithmic effects on the diversity of consumption on spotify.
\newblock In \emph{WWW}.

\bibitem[{Argyle et~al.(2023)Argyle, Busby, Fulda, Gubler, Rytting, and
  Wingate}]{argyle2023out}
Lisa~P Argyle, Ethan~C Busby, Nancy Fulda, Joshua~R Gubler, Christopher
  Rytting, and David Wingate. 2023.
\newblock Out of one, many: Using language models to simulate human samples.
\newblock \emph{Political Analysis}, 31(3):337--351.

\bibitem[{Chen(2023)}]{chen2023palr}
Zheng Chen. 2023.
\newblock Palr: Personalization aware llms for recommendation.
\newblock \emph{arXiv preprint arXiv:2305.07622}.

\bibitem[{Christakopoulou et~al.(2016)Christakopoulou, Radlinski, and
  Hofmann}]{christakopoulou2016towards}
Konstantina Christakopoulou, Filip Radlinski, and Katja Hofmann. 2016.
\newblock Towards conversational recommender systems.
\newblock In \emph{KDD}.

\bibitem[{Fan et~al.(2023)Fan, Zhao, Li, Liu, Mei, Wang, Tang, and
  Li}]{fan2023recommender}
Wenqi Fan, Zihuai Zhao, Jiatong Li, Yunqing Liu, Xiaowei Mei, Yiqi Wang,
  Jiliang Tang, and Qing Li. 2023.
\newblock Recommender systems in the era of large language models (llms).
\newblock \emph{arXiv preprint arXiv:2307.02046}.

\bibitem[{Gao et~al.(2023)Gao, Lan, Lu, Mao, Piao, Wang, Jin, and
  Li}]{gao2023s}
Chen Gao, Xiaochong Lan, Zhihong Lu, Jinzhu Mao, Jinghua Piao, Huandong Wang,
  Depeng Jin, and Yong Li. 2023.
\newblock S3: Social-network simulation system with large language
  model-empowered agents.
\newblock \emph{arXiv preprint arXiv:2307.14984}.

\bibitem[{Gao et~al.(2021)Gao, Lei, He, de~Rijke, and Chua}]{gao2021advances}
Chongming Gao, Wenqiang Lei, Xiangnan He, Maarten de~Rijke, and Tat-Seng Chua.
  2021.
\newblock Advances and challenges in conversational recommender systems: A
  survey.
\newblock \emph{AI Open}.

\bibitem[{Guo et~al.(2018)Guo, Wu, Cheng, Rennie, Tesauro, and
  Feris}]{guo2018dialog}
Xiaoxiao Guo, Hui Wu, Yu~Cheng, Steven Rennie, Gerald Tesauro, and Rogerio
  Feris. 2018.
\newblock Dialog-based interactive image retrieval.
\newblock \emph{NeurIPS}.

\bibitem[{H{\"a}m{\"a}l{\"a}inen et~al.(2023)H{\"a}m{\"a}l{\"a}inen, Tavast,
  and Kunnari}]{hamalainen2023evaluating}
Perttu H{\"a}m{\"a}l{\"a}inen, Mikke Tavast, and Anton Kunnari. 2023.
\newblock Evaluating large language models in generating synthetic hci research
  data: a case study.
\newblock In \emph{CHI}.

\bibitem[{Harper and Konstan(2015)}]{harper2015movielens}
F~Maxwell Harper and Joseph~A Konstan. 2015.
\newblock The movielens datasets: History and context.
\newblock \emph{ACM TIIS}.

\bibitem[{He et~al.(2023)He, Xie, Jha, Steck, Liang, Feng, Majumder, Kallus,
  and McAuley}]{he2023large}
Zhankui He, Zhouhang Xie, Rahul Jha, Harald Steck, Dawen Liang, Yesu Feng,
  Bodhisattwa~Prasad Majumder, Nathan Kallus, and Julian McAuley. 2023.
\newblock Large language models as zero-shot conversational recommenders.
\newblock In \emph{CIKM}.

\bibitem[{Hou et~al.(2023)Hou, Zhang, Lin, Lu, Xie, McAuley, and
  Zhao}]{hou2023large}
Yupeng Hou, Junjie Zhang, Zihan Lin, Hongyu Lu, Ruobing Xie, Julian McAuley,
  and Wayne~Xin Zhao. 2023.
\newblock Large language models are zero-shot rankers for recommender systems.
\newblock \emph{arXiv preprint arXiv:2305.08845}.

\bibitem[{Kang et~al.(2023)Kang, Ni, Mehta, Sathiamoorthy, Hong, Chi, and
  Cheng}]{kang2023llms}
Wang-Cheng Kang, Jianmo Ni, Nikhil Mehta, Maheswaran Sathiamoorthy, Lichan
  Hong, Ed~Chi, and Derek~Zhiyuan Cheng. 2023.
\newblock Do llms understand user preferences? evaluating llms on user rating
  prediction.
\newblock \emph{arXiv preprint arXiv:2305.06474}.

\bibitem[{Lei et~al.(2020{\natexlab{a}})Lei, He, Miao, Wu, Hong, Kan, and
  Chua}]{lei2020estimation}
Wenqiang Lei, Xiangnan He, Yisong Miao, Qingyun Wu, Richang Hong, Min-Yen Kan,
  and Tat-Seng Chua. 2020{\natexlab{a}}.
\newblock Estimation-action-reflection: Towards deep interaction between
  conversational and recommender systems.
\newblock In \emph{WSDM}.

\bibitem[{Lei et~al.(2020{\natexlab{b}})Lei, Zhang, He, Miao, Wang, Chen, and
  Chua}]{lei2020interactive}
Wenqiang Lei, Gangyi Zhang, Xiangnan He, Yisong Miao, Xiang Wang, Liang Chen,
  and Tat-Seng Chua. 2020{\natexlab{b}}.
\newblock Interactive path reasoning on graph for conversational
  recommendation.
\newblock In \emph{KDD}.

\bibitem[{Li et~al.(2023)Li, Zhang, Wang, Xiong, Lu, and
  Medioni}]{li2023gpt4rec}
Jinming Li, Wentao Zhang, Tian Wang, Guanglei Xiong, Alan Lu, and Gerard
  Medioni. 2023.
\newblock Gpt4rec: A generative framework for personalized recommendation and
  user interests interpretation.
\newblock \emph{arXiv preprint arXiv:2304.03879}.

\bibitem[{Li et~al.(2018)Li, Ebrahimi~Kahou, Schulz, Michalski, Charlin, and
  Pal}]{li2018towards}
Raymond Li, Samira Ebrahimi~Kahou, Hannes Schulz, Vincent Michalski, Laurent
  Charlin, and Chris Pal. 2018.
\newblock Towards deep conversational recommendations.
\newblock In \emph{NeurIPS}.

\bibitem[{Mikolov et~al.(2013)Mikolov, Chen, Corrado, and
  Dean}]{mikolov2013efficient}
Tomas Mikolov, Kai Chen, Greg Corrado, and Jeffrey Dean. 2013.
\newblock Efficient estimation of word representations in vector space.
\newblock \emph{arXiv preprint arXiv:1301.3781}.

\bibitem[{Momennejad et~al.(2023)Momennejad, Hasanbeig, Vieira, Sharma, Ness,
  Jojic, Palangi, and Larson}]{momennejad2023evaluating}
Ida Momennejad, Hosein Hasanbeig, Felipe Vieira, Hiteshi Sharma, Robert~Osazuwa
  Ness, Nebojsa Jojic, Hamid Palangi, and Jonathan Larson. 2023.
\newblock Evaluating cognitive maps and planning in large language models with
  cogeval.
\newblock \emph{arXiv preprint arXiv:2309.15129}.

\bibitem[{Moon et~al.(2019)Moon, Shah, Kumar, and Subba}]{moon2019opendialkg}
Seungwhan Moon, Pararth Shah, Anuj Kumar, and Rajen Subba. 2019.
\newblock Opendialkg: Explainable conversational reasoning with attention-based
  walks over knowledge graphs.
\newblock In \emph{ACL}.

\bibitem[{OpenAI(2021)}]{openai2021}
OpenAI. 2021.
\newblock \href {http://www.openai.com/about/} {About openai}.

\bibitem[{Owoicho et~al.(2023)Owoicho, Sekulic, Aliannejadi, Dalton, and
  Crestani}]{owoicho2023exploiting}
Paul Owoicho, Ivan Sekulic, Mohammad Aliannejadi, Jeffrey Dalton, and Fabio
  Crestani. 2023.
\newblock Exploiting simulated user feedback for conversational search:
  Ranking, rewriting, and beyond.
\newblock In \emph{SIGIR}.

\bibitem[{Pan et~al.(2023)Pan, Chen, Xu, Che, and Qin}]{pan2023preliminary}
Wenbo Pan, Qiguang Chen, Xiao Xu, Wanxiang Che, and Libo Qin. 2023.
\newblock A preliminary evaluation of chatgpt for zero-shot dialogue
  understanding.
\newblock \emph{arXiv preprint arXiv:2304.04256}.

\bibitem[{Park et~al.(2023)Park, O'Brien, Cai, Morris, Liang, and
  Bernstein}]{park2023generative}
Joon~Sung Park, Joseph~C O'Brien, Carrie~J Cai, Meredith~Ringel Morris, Percy
  Liang, and Michael~S Bernstein. 2023.
\newblock Generative agents: Interactive simulacra of human behavior.
\newblock In \emph{UIST}.

\bibitem[{Qian et~al.(2023)Qian, Cong, Yang, Chen, Su, Xu, Liu, and
  Sun}]{qian2023communicative}
Chen Qian, Xin Cong, Cheng Yang, Weize Chen, Yusheng Su, Juyuan Xu, Zhiyuan
  Liu, and Maosong Sun. 2023.
\newblock Communicative agents for software development.
\newblock \emph{arXiv preprint arXiv:2307.07924}.

\bibitem[{Reimers and Gurevych(2019)}]{reimers2019sentence}
Nils Reimers and Iryna Gurevych. 2019.
\newblock Sentence-bert: Sentence embeddings using siamese bert-networks.
\newblock \emph{EMNLP}.

\bibitem[{Santurkar et~al.(2023)Santurkar, Durmus, Ladhak, Lee, Liang, and
  Hashimoto}]{santurkar2023whose}
Shibani Santurkar, Esin Durmus, Faisal Ladhak, Cinoo Lee, Percy Liang, and
  Tatsunori Hashimoto. 2023.
\newblock Whose opinions do language models reflect?
\newblock In \emph{ICML}.

\bibitem[{Sun and Zhang(2018)}]{sun2018conversational}
Yueming Sun and Yi~Zhang. 2018.
\newblock Conversational recommender system.
\newblock In \emph{SIGIR}.

\bibitem[{Wang et~al.(2023{\natexlab{a}})Wang, Zhang, Chen, Lin, Song, Zhao,
  and Wen}]{wang2023recagent}
Lei Wang, Jingsen Zhang, Xu~Chen, Yankai Lin, Ruihua Song, Wayne~Xin Zhao, and
  Ji-Rong Wen. 2023{\natexlab{a}}.
\newblock Recagent: A novel simulation paradigm for recommender systems.
\newblock \emph{arXiv preprint arXiv:2306.02552}.

\bibitem[{Wang et~al.(2023{\natexlab{b}})Wang, Tang, Zhao, Wang, and
  Wen}]{wang2023rethinking}
Xiaolei Wang, Xinyu Tang, Wayne~Xin Zhao, Jingyuan Wang, and Ji-Rong Wen.
  2023{\natexlab{b}}.
\newblock Rethinking the evaluation for conversational recommendation in the
  era of large language models.
\newblock In \emph{EMNLP}.

\bibitem[{Wang et~al.(2023{\natexlab{c}})Wang, Xu, Ai, and
  Srikumar}]{wang2023depth}
Zhenduo Wang, Zhichao Xu, Qingyao Ai, and Vivek Srikumar. 2023{\natexlab{c}}.
\newblock An in-depth investigation of user response simulation for
  conversational search.
\newblock \emph{arXiv preprint arXiv:2304.07944}.

\bibitem[{Xia et~al.(2023)Xia, Wu, Yu, Kim, Rossi, and Li}]{xia2023user}
Yu~Xia, Junda Wu, Tong Yu, Sungchul Kim, Ryan~A Rossi, and Shuai Li. 2023.
\newblock User-regulation deconfounded conversational recommender system with
  bandit feedback.
\newblock In \emph{KDD}.

\bibitem[{Yang et~al.(2023)Yang, Zhang, and Li}]{yang2023pyabsa}
Heng Yang, Chen Zhang, and Ke~Li. 2023.
\newblock Pyabsa: A modularized framework for reproducible aspect-based
  sentiment analysis.
\newblock In \emph{CIKM}.

\bibitem[{Zhang et~al.(2023)Zhang, Hou, Xie, Sun, McAuley, Zhao, Lin, and
  Wen}]{zhang2023agentcf}
Junjie Zhang, Yupeng Hou, Ruobing Xie, Wenqi Sun, Julian McAuley, Wayne~Xin
  Zhao, Leyu Lin, and Ji-Rong Wen. 2023.
\newblock Agentcf: Collaborative learning with autonomous language agents for
  recommender systems.
\newblock \emph{arXiv preprint arXiv:2310.09233}.

\bibitem[{Zhang and Balog(2020)}]{zhang2020evaluating}
Shuo Zhang and Krisztian Balog. 2020.
\newblock Evaluating conversational recommender systems via user simulation.
\newblock In \emph{KDD}.

\bibitem[{Zhang et~al.(2022)Zhang, Wang, and Balog}]{zhang2022analyzing}
Shuo Zhang, Mu-Chun Wang, and Krisztian Balog. 2022.
\newblock Analyzing and simulating user utterance reformulation in
  conversational recommender systems.
\newblock In \emph{SIGIR}.

\bibitem[{Zhao et~al.(2023)Zhao, Zhao, Lu, Wang, Tong, and
  Qin}]{zhao2023chatgpt}
Weixiang Zhao, Yanyan Zhao, Xin Lu, Shilong Wang, Yanpeng Tong, and Bing Qin.
  2023.
\newblock Is chatgpt equipped with emotional dialogue capabilities?
\newblock \emph{arXiv preprint arXiv:2304.09582}.

\end{thebibliography}
\bibliographystyle{acl_natbib}

\appendix

\section{Appendix}
\label{sec:appendix}

\subsection{Dataset statistics}
\label{sec:dataset_statistics}

The \textbf{ReDial} dataset contains $11,348$ conversations and $6,925$ movies.
The dataset was collected until 2018, and hence, movies up to this year are mentioned.
$1,309$ movies are used for \textit{\taskone}.

The original \textbf{Reddit} dataset contains $634,392$ conversations, $1,669,720$ turns, $36,247$ users, and $5,1203$ movies. 
We process this dataset in the following way: remove posts after 2021, remove comments without movie mentions, remove requests that are not about movies, and sample one head comment for each request.
The resulting dataset has $23,167$ requests, each with one comment. 
$9,974$ movies are used for \textit{\taskone}.

The \textbf{MovieLens} dataset contains ratings of $62,000$ movies rated by $162,000$ users. 
We sample $200$ movies rated by more than $5000$ users (frequent movies) and $200$ movies rated by less than $500$ and more than $50$ users (infrequent movies). 
While we also sample $300$ movies without considering rating frequency, we observe that due to the long-tail distribution of frequencies, this random sampling result in movies with significantly lower frequencies (median: 5, mode: 1 appears 49/300 times).
We nonetheless show the results.

The \textbf{IMDB} dataset originally consists of $1,083$ users and $22,918$ reviews.
Each of $928$ users has at least $11$ movie reviews.
$8,138$ movies are used for \textit{\taskone}.

\subsection{Prompts}\label{sec:prompts}

Here we provide the full prompts for all the tasks.

\begin{figure}[t]
    \centering
    \begin{subfigure}[b]{0.9\linewidth}
        \includegraphics[width=\linewidth]{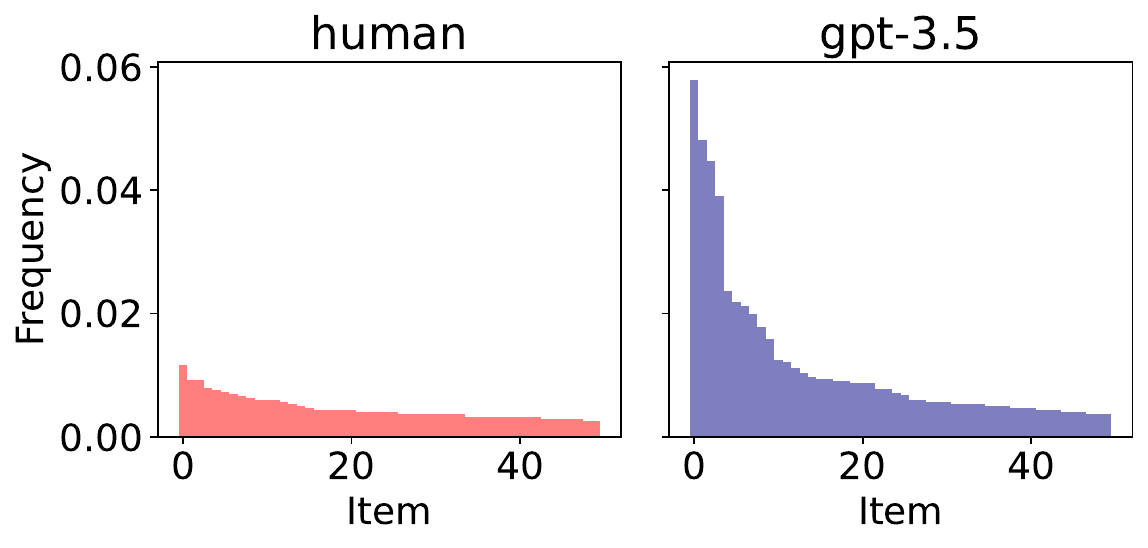}
    \end{subfigure}
    \begin{subfigure}[b]{0.9\linewidth}
        \includegraphics[width=\linewidth]{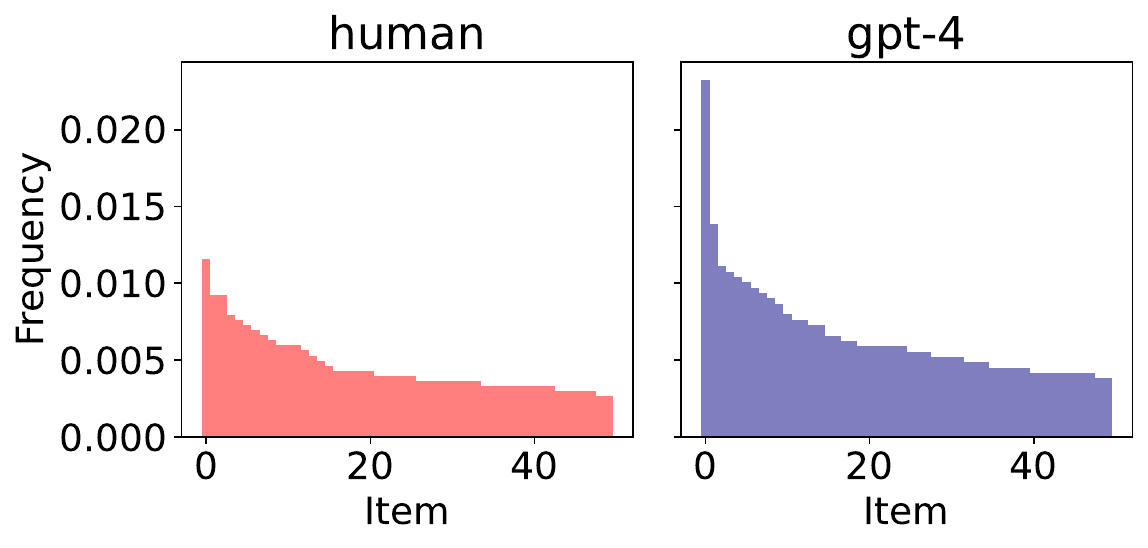}
    \end{subfigure}
    \begin{subfigure}[b]{0.9\linewidth}
        \includegraphics[width=\linewidth]{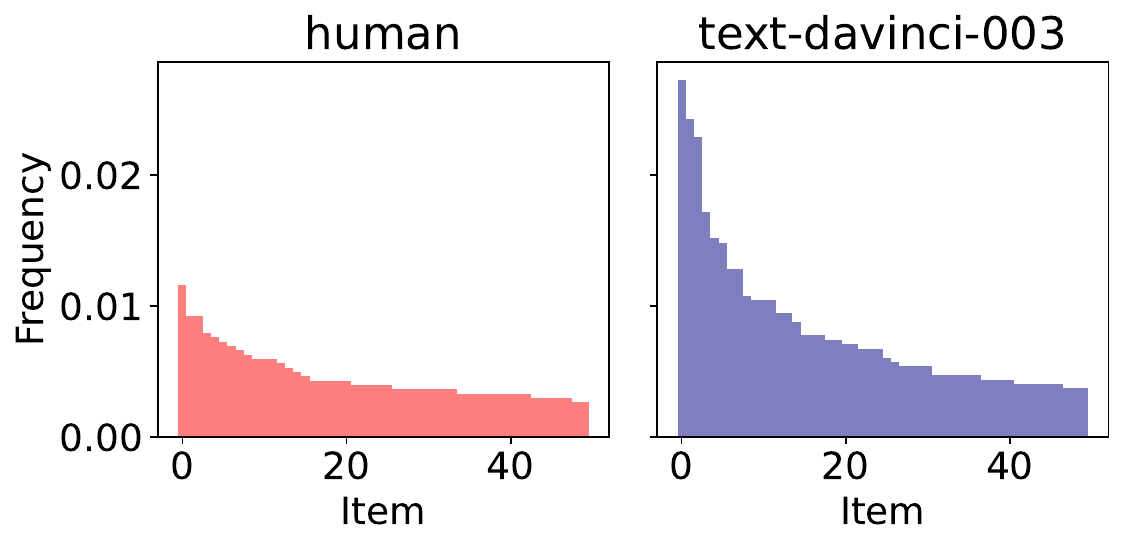}
    \end{subfigure}
    \caption{{\taskone} results for ReDial. Simulators are prompted with interaction history per ReDial conversation.}
    \label{fig:t1_redial}
\end{figure}
\begin{figure}[t]
    \centering
    \begin{subfigure}[b]{0.9\linewidth}
        \includegraphics[width=\linewidth]{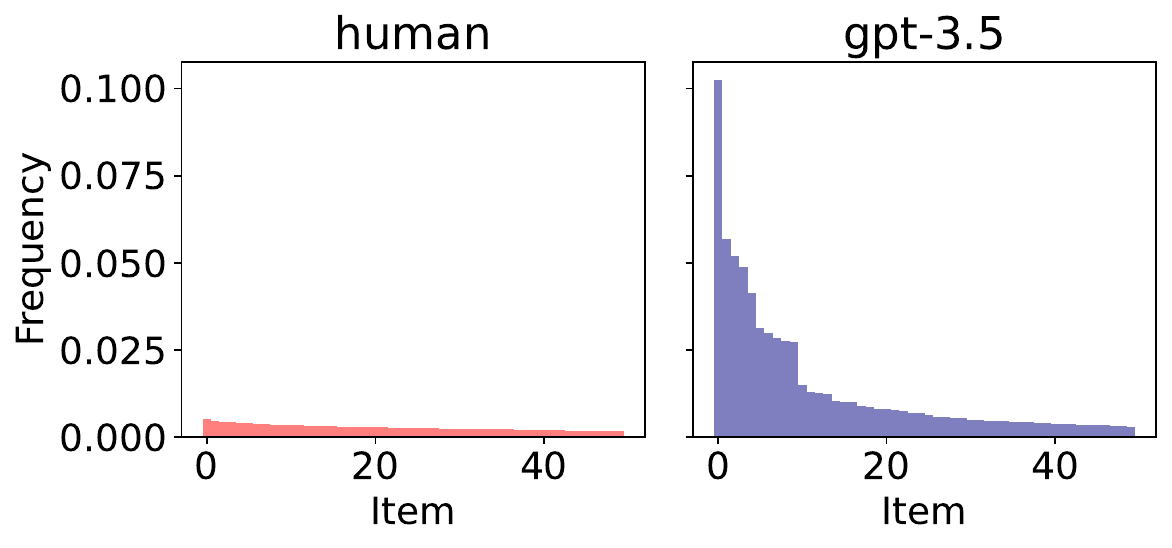}
    \end{subfigure}
    \begin{subfigure}[b]{0.9\linewidth}
        \includegraphics[width=\linewidth]{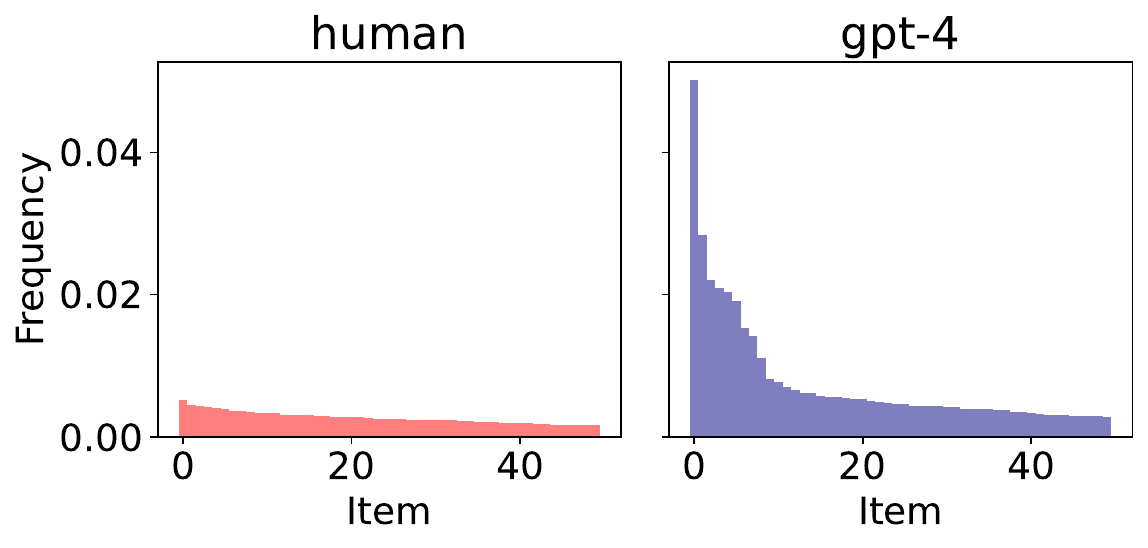}
    \end{subfigure}
    \begin{subfigure}[b]{0.9\linewidth}
        \includegraphics[width=\linewidth]{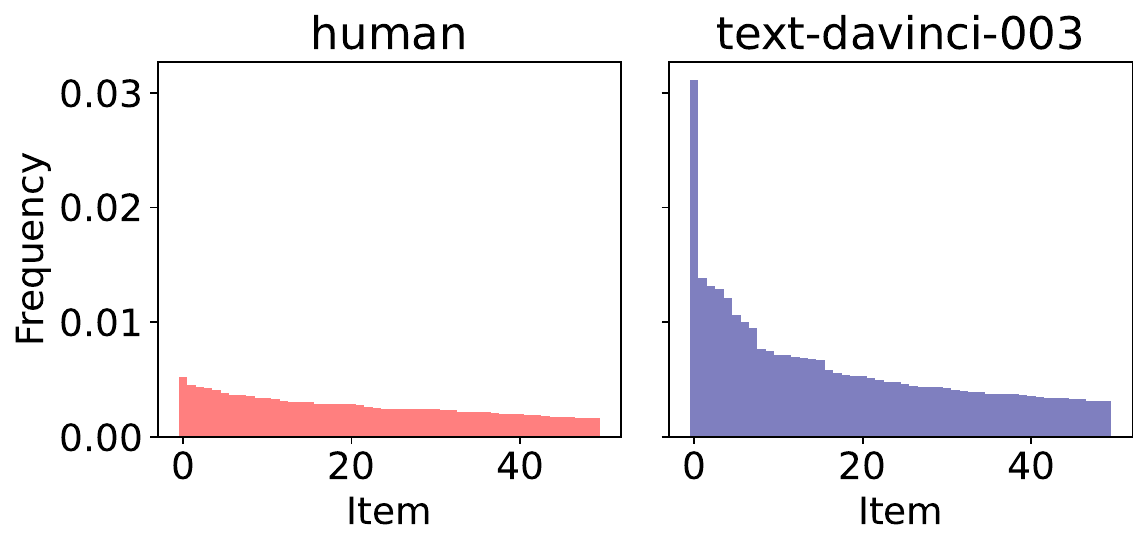}
    \end{subfigure}
    \caption{{\taskone} results for Reddit. Simulators are prompted with interaction history per Reddit request.}
    \label{fig:t1_reddit}
\end{figure}
\begin{figure}[t]
    \centering
    \begin{subfigure}[b]{0.9\linewidth}
        \includegraphics[width=\linewidth]{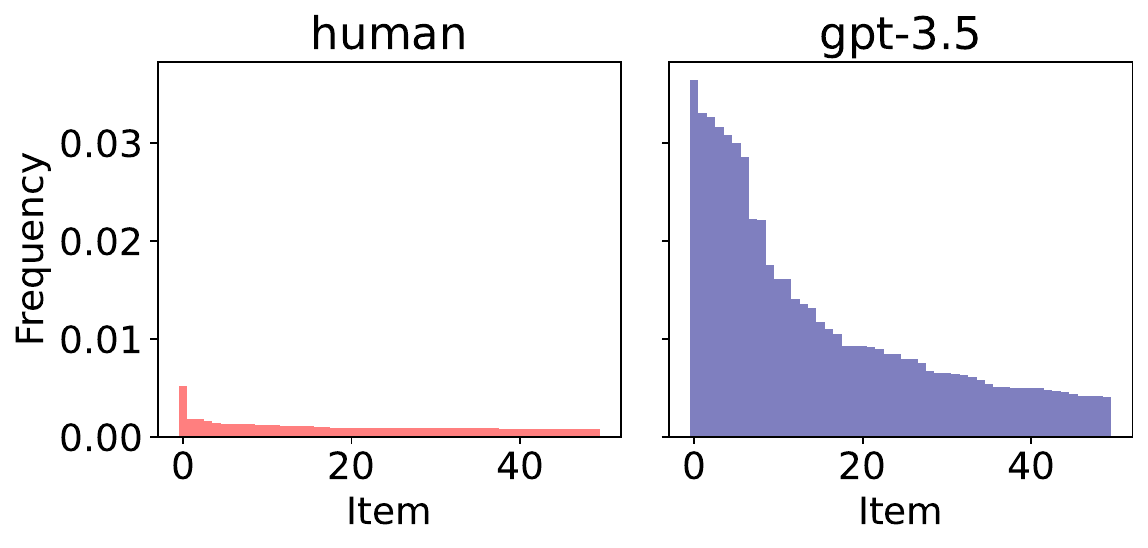}
    \end{subfigure}
    \begin{subfigure}[b]{0.9\linewidth}
        \includegraphics[width=\linewidth]{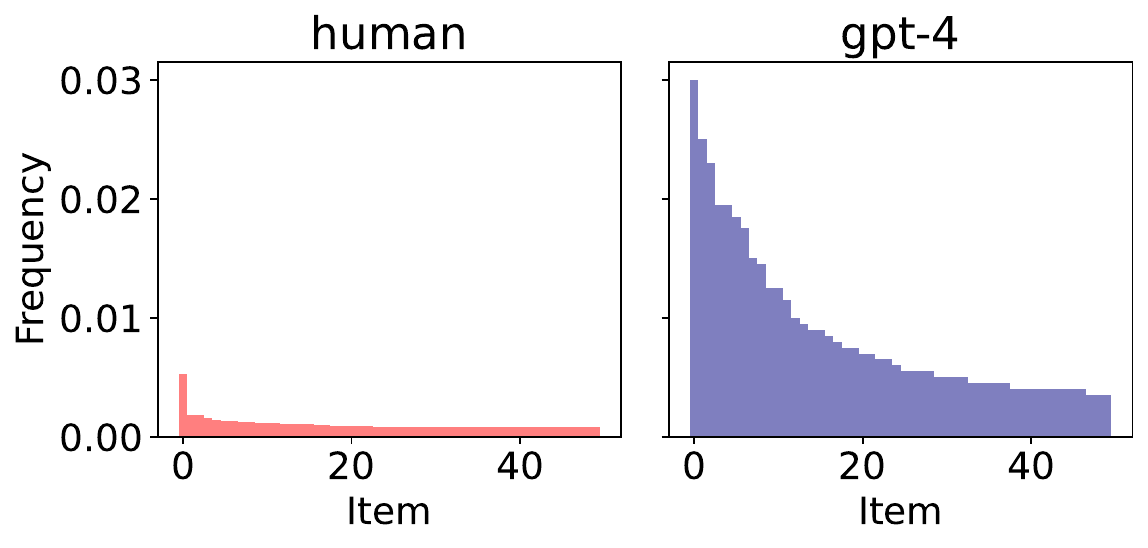}
    \end{subfigure}
    \begin{subfigure}[b]{0.9\linewidth}
        \includegraphics[width=\linewidth]{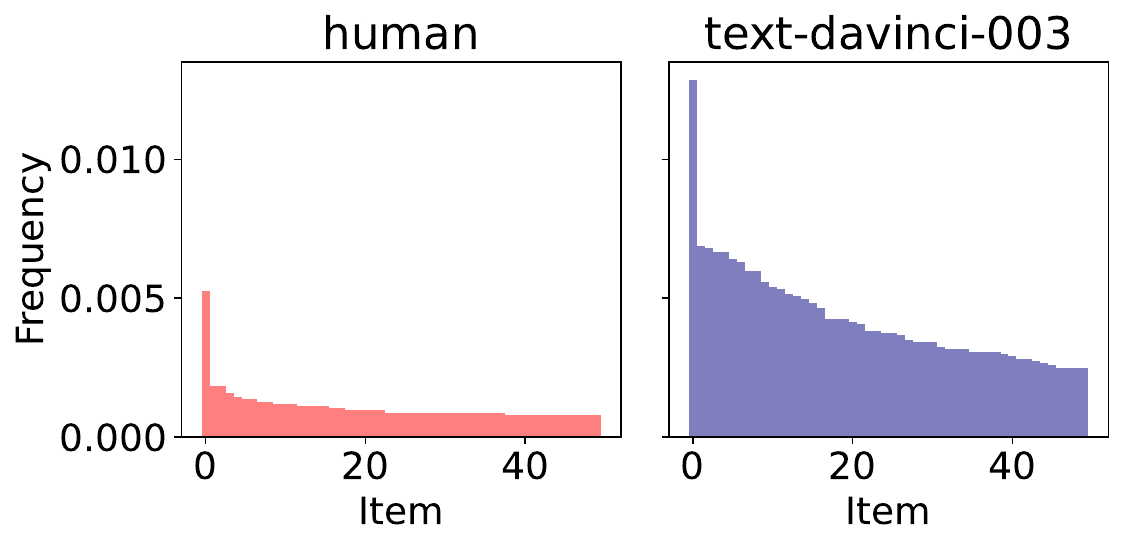}
    \end{subfigure}
    \caption{{\taskone} results for IMDB. Simulators are prompted with interaction history per IMDB user.}
    \label{fig:t1_imdb}
\end{figure}

\subsubsection{\taskone}

For the demographic information (\demographic) simulator, we randomly sample a prefix and a surname and ask to generate the movies one wants to talk about as this person.
\begin{quote}
Pretend to be \{prefix\} \{surname\}. You decide to talk about \{target\_num\} movies. What would these \{target\_num\} movies be? Reply as a list of <Title (yyyy)>. Say nothing else.    
\end{quote}

The interaction history (\interaction) simulator has slight variations according to the dataset where the interaction histories are sampled. 

IMDB: 10 movies and the corresponding review titles from each user.
\begin{quote}
A person leaves the following remarks on movies…\\
\{movie 1\}: \{review title 1\} \\
…\\
\{movie N\}: \{review title N\} \\
and proceeds to talk about \{target\_num\} more movies. What would these \{target\_num\} movies be? Reply as a list of <Title (yyyy)>. Say nothing else.
\end{quote}

Reddit: one movie and the UTC time it was mentioned, from each request.
\begin{quote}
At UTC time \{time\}, a person starts to talk about the movies \{movies\} and proceeds to talk about \{target\_num\} more. What would these \{target\_num\} movies be? Reply as a list of <Title (yyyy)>. Say nothing else.
\end{quote}

ReDial: 2 movies from the seeker side of each conversation.
\begin{quote}
A person mentions \{movies\} in a conversation about movies and proceeds to mention \{target\_num\} more. What would these \{target\_num\} movies be? Reply as a list of <Title (yyyy)>. Say nothing else.
\end{quote}

Target number of items is the number of total movies in the entry, minus the number of movies used as interaction history.

\subsubsection{\tasktwo}

We randomly sample a prefix and a surname for the demographic information (\demographic) simulator.

\begin{quote}
Pretend to be \{prefix\} \{surname\}. You watched the movie \{movie\}. Did you like the movie? Answer Yes or No. Don't say anything else.
\end{quote}

We add a pickiness trait (\pickiness) by randomly selecting among the three levels of pickiness: extremely picky, moderately picky, and not picky.
\begin{quote}
Pretend to be \{prefix\} \{surname\}. You are \{pickiness\} about movies. You watched the movie \{movie\}. Did you like the movie? Answer Yes or No. Don't say anything else.
\end{quote}

\subsubsection{\taskthree}

Similar to \tasktwo, we prompt the {\demographic} simulator
\begin{quote}
Pretend to be \{prefix\} \{surname\}. You watched the movie \{movie\}. What are your thoughts on this movie? Answer should not exceed \{review\_len\} characters.
\end{quote}
and the {\pickiness} simulator
\begin{quote}
Pretend to be \{prefix\} \{surname\}. You are \{pickiness\} about movies. You watched the movie \{movie\}. What are your thoughts on this movie? Answer should not exceed \{review\_len\} characters.
\end{quote}

Target review length is determined by the review length in the processed IMDB baseline, where one review is sampled per user.

\subsubsection{\taskfour}

From each Reddit request, we use the movie names mentioned and the length of the request to prompt the following.

\begin{quote}
Generate a movie recommendation request. Include (but do not request) the following movies in your text: \{movies\}. Make sure the length of the request is approximately \{target\_len\} characters.
\end{quote}

\subsubsection{\taskfive}

For the accept/reject task, we sample one positive and one negative response (recommendation) per request and use the following prompt for each pair.
\begin{quote}
In the following conversation, a USER asks for movie recommendations. Your task is to act like the USER by giving the following responses to the AGENT's recommendation:
If the recommendation is coherent to your request, answer Accept.
If the recommendation is incoherent to your request, answer Reject.
Simply answer Accept or Reject.\\
USER: \{request\} \\
AGENT: \{response\} \\
USER (answer Accept or Reject):
\end{quote}

For the comparison task, we use both the positive and negative responses of a request in a single prompt. 
To eliminate position bias, we randomly choose assign the responses to the AGENTs.
That is, the positive response is assigned to either AGENT 1 or AGENT 2 with equal probability.
\begin{quote}
A USER asks for movie recommendation. AGENT 1 and AGENT 2 gives recommendations. Your task is to choose the AGENT that gives better recommendations. Simply answer AGENT 1 or AGENT 2. You HAVE to choose one.\\
USER: \{request\} \\
AGENT 1's response: \{response\} \\
AGENT 2's response: \{response\} \\
Which response is better? \\
(Reply AGENT 1 or AGENT 2)
\end{quote}

\begin{figure}[t]
    \centering
    \includegraphics[width=0.5\textwidth]{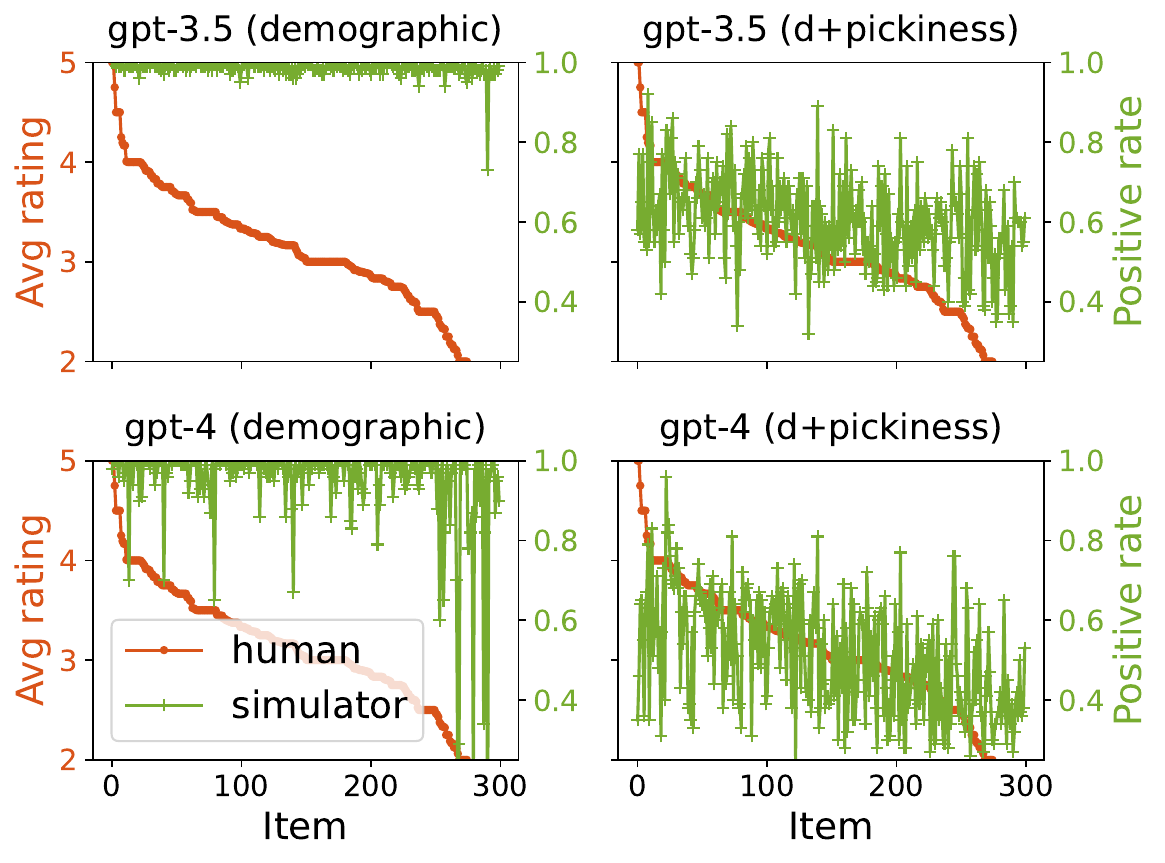}
    \caption{Preference trends of humans and simulators. See Table~\ref{tab:correlation-all-300} for correlation coefficients.}
    \label{fig:all-300}
\end{figure}

\begin{table}[t] 
\centering
\caption{Correlation coefficient between human and simulator preferences. Movies are randomly sampled across all frequency levels, resulting in a sample with movies with very low frequencies.}
\label{tab:correlation-all-300}
\scalebox{0.9}{
\begin{tabular}{c|c}
\hline
\multicolumn{2}{c}{Demographic information} \\ 
\hline
gpt-3.5 & 0.27  \\ 
gpt-4 &  0.29 \\
\hline
\multicolumn{2}{c}{Demographic information + Pickiness} \\ 
\hline
gpt-3.5 & 0.32 \\
gpt-4 & 0.44 \\
\hline
\end{tabular}
}
\end{table}

For cases ($20$ cases for each configuration) where we ask to provide reasons, we add the following to the prompt.
\begin{quote}
Provide a short reason (less than 40 words) for your response.
\end{quote}

\begin{figure*}[t!]
    \centering
    \begin{subfigure}[b]{0.3\linewidth}
        \includegraphics[width=\linewidth]{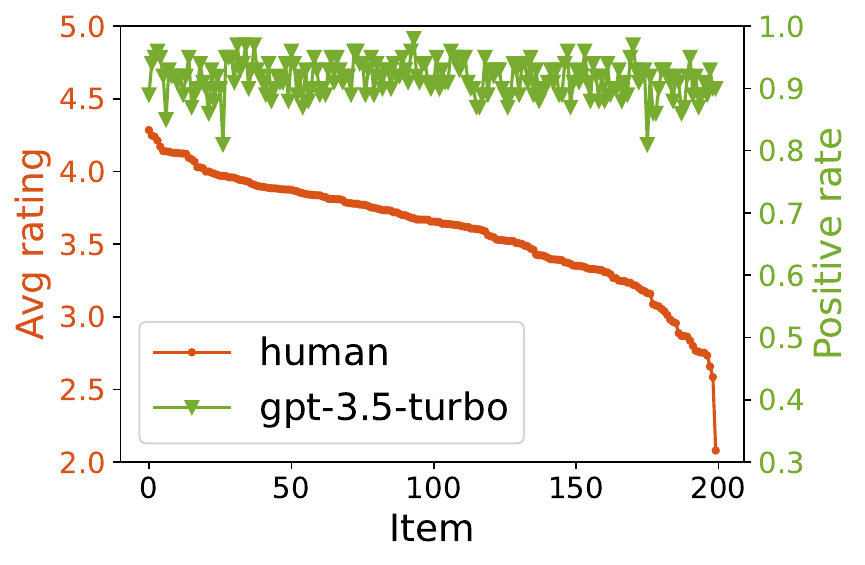}
        \caption{gpt-3.5 + \demographic}
    \end{subfigure}
    \hfill 
    \begin{subfigure}[b]{0.3\linewidth}
        \includegraphics[width=\linewidth]{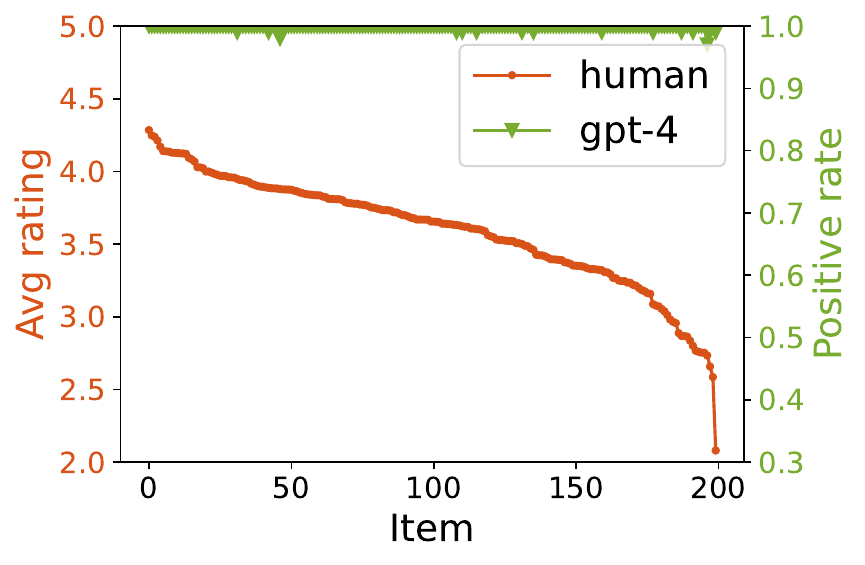}
        \caption{gpt-4 + \demographic}
    \end{subfigure}
    \hfill 
    \begin{subfigure}[b]{0.3\linewidth}
        \includegraphics[width=\linewidth]{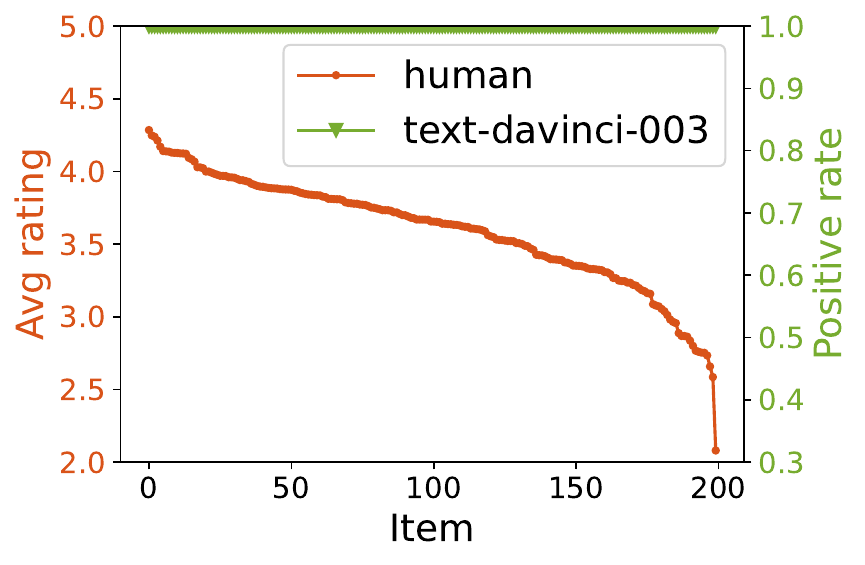}
        \caption{text-davinci + \demographic}
    \end{subfigure}
    
    \begin{subfigure}[b]{0.3\linewidth}
        \includegraphics[width=\linewidth]{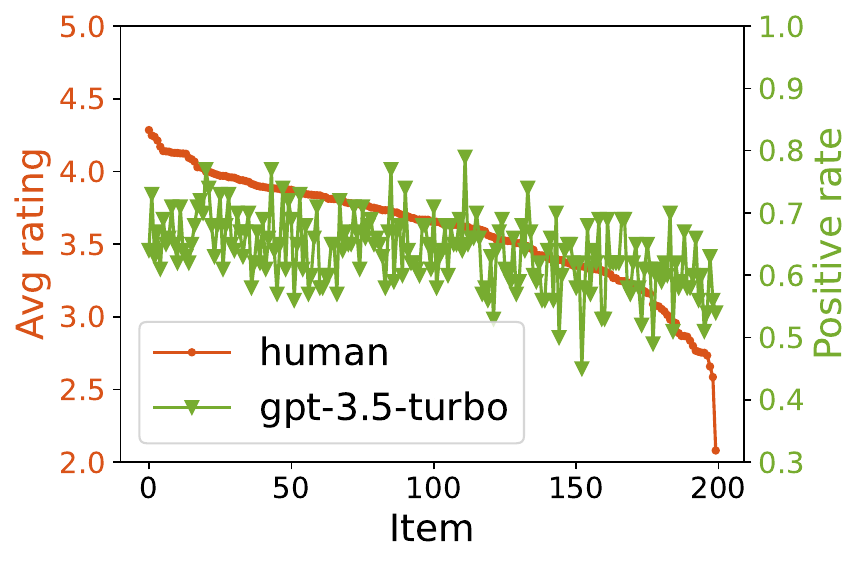}
        \caption{gpt-3.5 + \pickiness}
    \end{subfigure}
    \hfill
    \begin{subfigure}[b]{0.3\linewidth}
        \includegraphics[width=\linewidth]{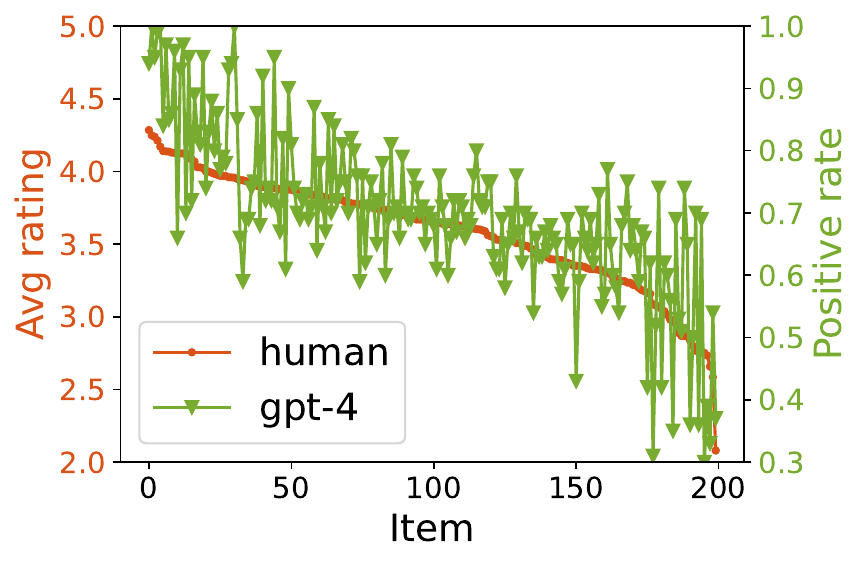}
        \caption{gpt-4 + \pickiness}
    \end{subfigure}
    \hfill
    \begin{subfigure}[b]{0.3\linewidth}
        \includegraphics[width=\linewidth]{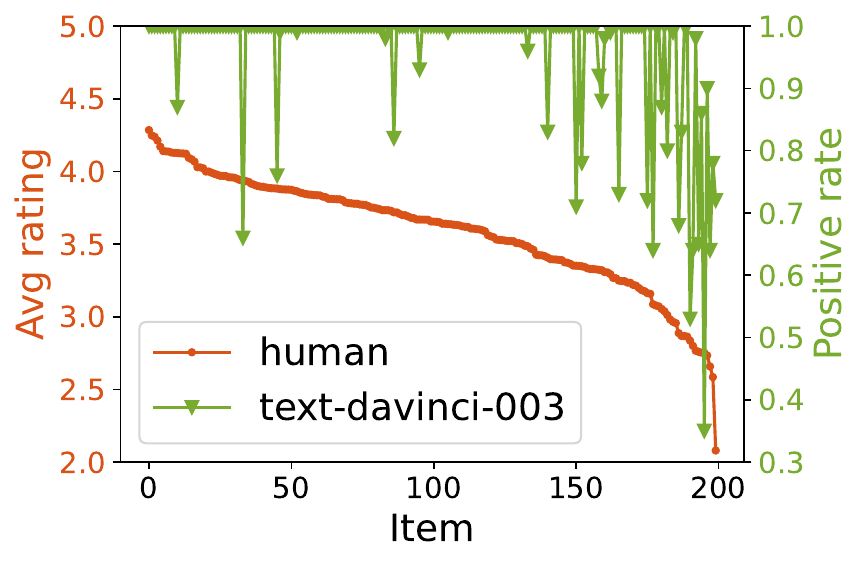}
        \caption{text-davinci + \pickiness}
    \end{subfigure}
    \caption{Preference trends of frequent movies}
    \label{fig:frequent}
\end{figure*}

\begin{figure*}[t!]
    \centering
    \begin{subfigure}[b]{0.3\linewidth}
        \includegraphics[width=\linewidth]{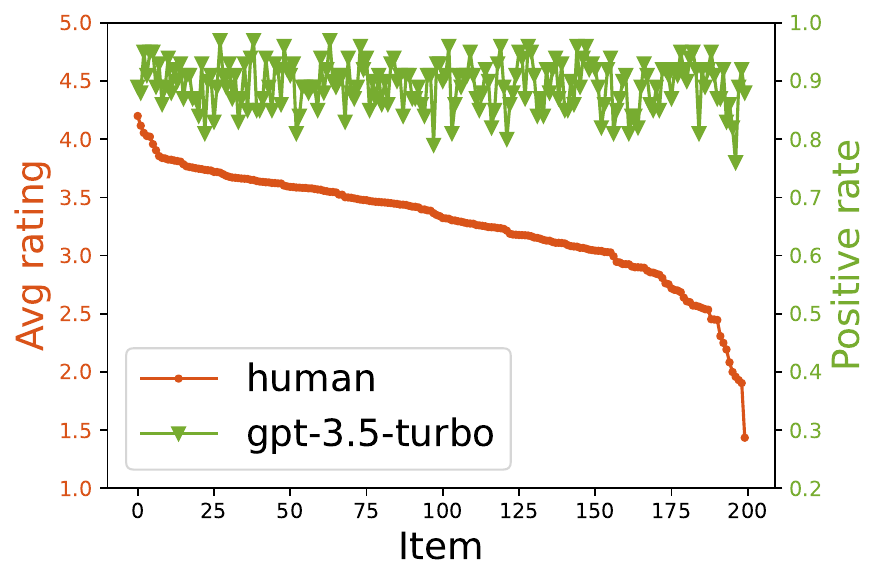}
        \caption{gpt-3.5 + \demographic}
    \end{subfigure}
    \hfill 
    \begin{subfigure}[b]{0.3\linewidth}
        \includegraphics[width=\linewidth]{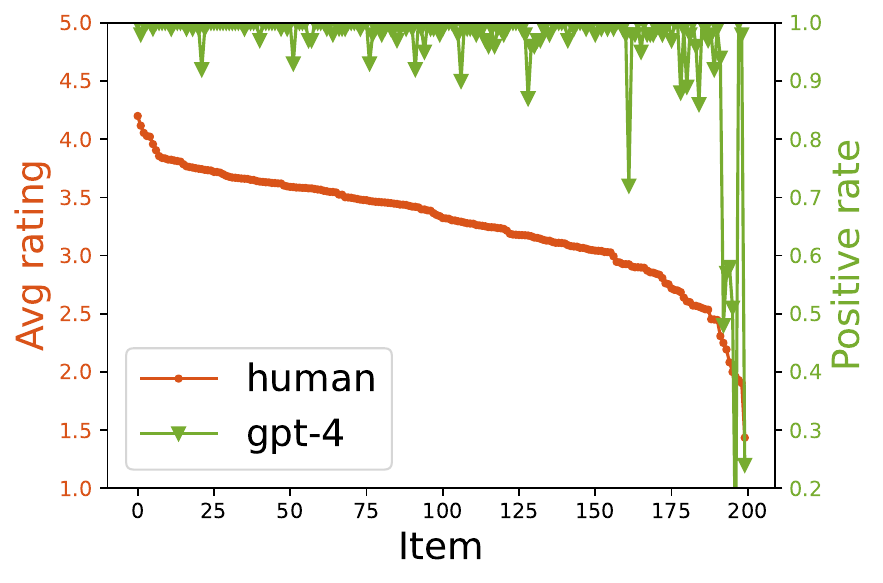}
        \caption{gpt-4 + \demographic}
    \end{subfigure}
    \hfill 
    \begin{subfigure}[b]{0.3\linewidth}
        \includegraphics[width=\linewidth]{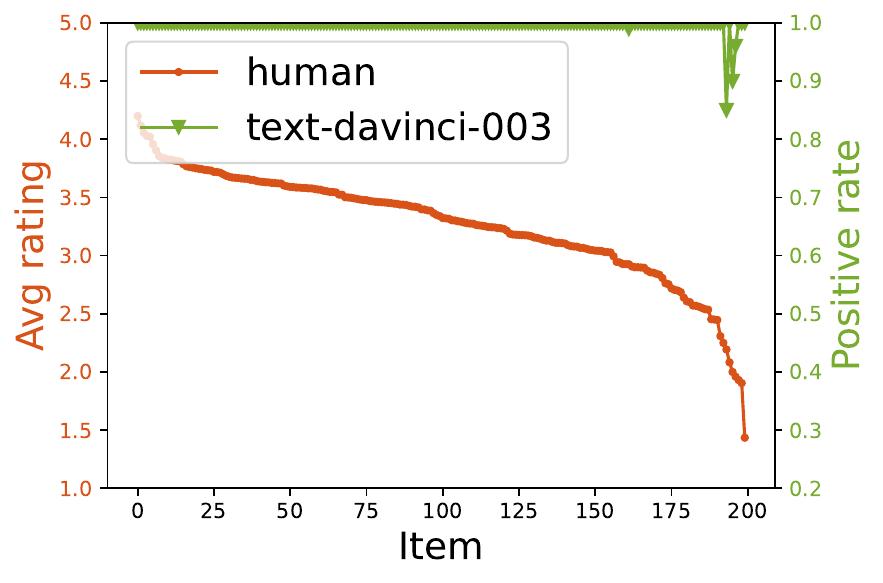}
        \caption{text-davinci + \demographic}
    \end{subfigure}
    
    \begin{subfigure}[b]{0.3\linewidth}
        \includegraphics[width=\linewidth]{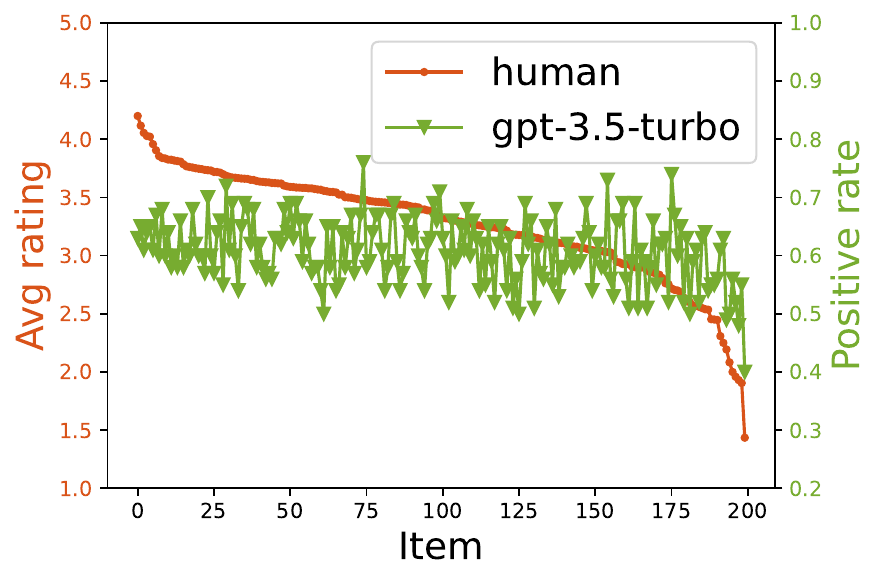}
        \caption{gpt-3.5 + \pickiness}
    \end{subfigure}
    \hfill
    \begin{subfigure}[b]{0.3\linewidth}
        \includegraphics[width=\linewidth]{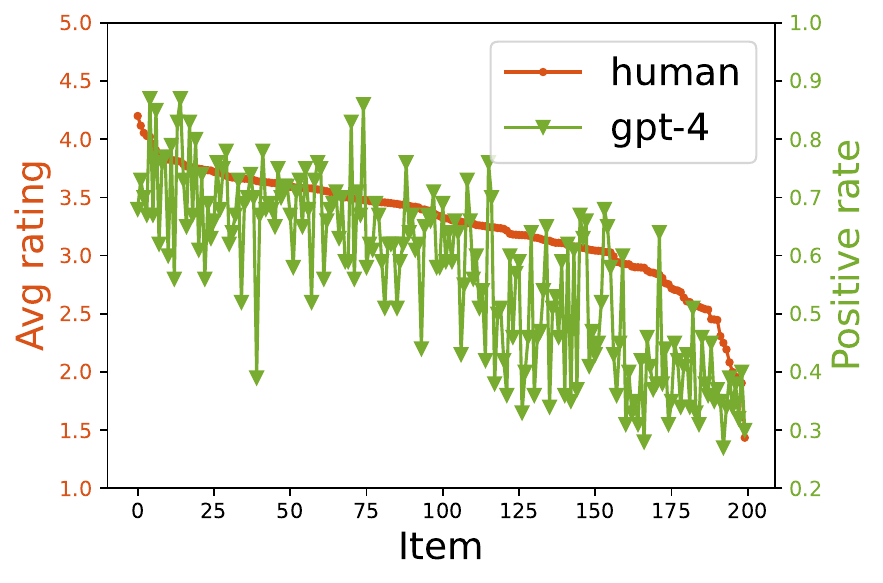}
        \caption{gpt-4 + \pickiness}
    \end{subfigure}
    \hfill
    \begin{subfigure}[b]{0.3\linewidth}
        \includegraphics[width=\linewidth]{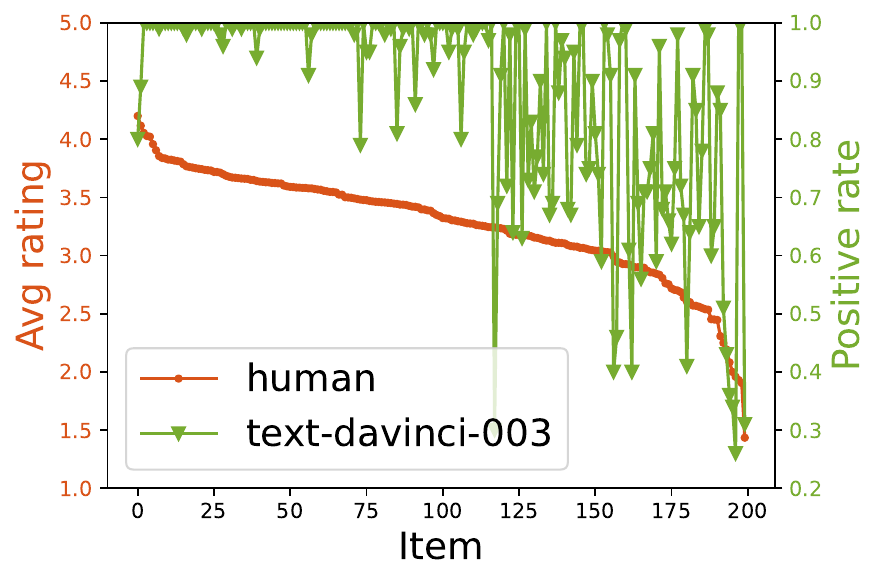}
        \caption{text-davinci + \pickiness}
    \end{subfigure}
    \caption{Preference trends of infrequent movies}
    \label{fig:infrequent}
\end{figure*}

\subsection{More results}\label{sec:more_results}

\subsubsection{Results from \taskone}
\label{sec:more_taskone}

We compare the distributions of items mentioned by simulators and real users. 
We plot human and simulator distributions side-by-side per dataset:
ReDial (Figure~\ref{fig:t1_redial}), Reddit (Figure~\ref{fig:t1_reddit}), and IMDB (Figure~\ref{fig:t1_imdb}).
Note that the three human distributions within each figure is the same, but scaled differently according to the scale of different language models.

\subsubsection{Results from \tasktwo}
\label{sec:more_tasktwo}

We compare the trends of average rating and positive rate (proportion of simulators that answered that they liked the movie). 
Results for items randomly sampled regardless of frequency levels are shown in Table~\ref{tab:correlation-all-300} and Figure~\ref{fig:all-300}.\footnote{These experiments were conducted after we received our paper reviews. OpenAI's configurations might have been updated or modified in the meantime. We did not obtain results for text-davinci-003, since by the time we ran these experiments, the model was deprecated.}
However, these movies often have very low frequency values (median: 5, mode: 1 appears 49/300 times); the average ratings of these movies may not reflect true user preferences. 
Plots for frequent items are shown in Figure~\ref{fig:frequent} and infrequent items in Figure~\ref{fig:infrequent}.

\subsubsection{More {\taskfive} examples}
\label{sec:more_taskfive}
We provide more examples generated by simulators, when asked to provide a reason for their feedback. 
All examples are from gpt-3.5-turbo.

In our first example, The human user asks for movies, and possibly even shows. 
However, the simulator rejects the relevant recommendation, saying that it is a show not a movie:

\begin{itemize}
    \item \textbf{Request:} Looking for a Murder Mystery Serial Killer movie. Recently watched the Bone Collector, Taking Lives and Copy Cat, if you have any recommendations, (even ones that were just fine, or even sub-par) I'd love to see them! Or even a show like Criminal Minds, where each episode is it's own little movie.
    \item \textbf{Positive Rec:} Not movie, but show called : La Mante
    \item \textbf{Feedback:} Reject. La Mante is recommended as a show, not a movie.
\end{itemize}

In another example, a list of relevant movies is recommended, while including the one that the human user has already watched.
The simulator rejects the recommendation just because it includes this movie.
(We alter certain words that could be perceived as offensive, although their original usage was intended positively.)
\begin{itemize}
    \item \textbf{Request:} What are the most disturbing up movies on UK Netflix right now!. In the past there have been some great films on UK Netflix to mess with you, (antichrist, I saw the devil, the original oldboy, Ichi the killer, the human centipede, requiem for a dream etc.) But it seems there aren't any good ones to find that I haven't seen and that are still streaming! Most of these have been taken off now and I'm after something new but along similar lines... does anyone have any suggestions? Thanks in advance!
    \item \textbf{Positive Rec:} It Follows (2014), Victoria (2015), Antichrist (2009)
    \item \textbf{Feedback:} Reject. Reason: Already mentioned and seen Antichrist in the past. Looking for something new but along similar lines.
\end{itemize}

Finally, a user wants to be cheered up, wanting to be inspired for a better future. 
The simulator understood the mission, but chooses the negative recommendation (\textit{Love (2005)}, a romance movie) over the positive one (\textit{Trolls (2016)}, an animation about happiness).
The reason is unclear.

\begin{itemize}
    \item \textbf{Request:} Movie about positively changing in the future. Hi, I got into some trouble recently with the law and in general have been reckless in my drug use. I fear the social and legal reprecussions. I'd like to watch something that reminds me good times can still be had in the future. I'd just like something to cheer me up, no comedies please.
    \item \textbf{Positive Rec:} Trolls (2016)
    \item \textbf{Negative Rec:} Love (2005)
    \item \textbf{Feedback:} Love (2005). Love (2005) seems like a better recommendation as it may provide the user with a positive outlook on the future and remind them that good times can still be experienced even in difficult situations.
\end{itemize}

\end{document}